\documentclass[10pt,journal,compsoc]{IEEEtran}

\ifCLASSOPTIONcompsoc
  \usepackage[nocompress]{cite}
\else
  \usepackage{cite}
\fi

\ifCLASSINFOpdf
\else
\fi

\usepackage{enumitem}
\usepackage{bm}
\usepackage{amsmath}
\usepackage{amsthm}
\usepackage{amssymb}
\usepackage{graphicx}
\usepackage[linesnumbered,ruled,vlined]{algorithm2e}
\usepackage{float}
\usepackage{booktabs}
\usepackage{subfigure}
\usepackage{makecell}
\usepackage{multirow}
\usepackage{lineno}
\usepackage{siunitx}
\usepackage{url}
\usepackage{dsfont}
\usepackage{xcolor} 
\newtheorem{definition}{Definition}
\newcommand{\citet}{\cite}

\newcommand{\figref}[1]{Fig. \ref{#1}}
\newcommand{\tabref}[1]{Table \ref{#1}}
\newcommand{\secref}[1]{Section \ref{#1}}
\newcommand{\equref}[1]{Equation (\ref{#1})}
\newcommand{\algoref}[1]{Algorithm \ref{#1}}
\newcommand{\myref}[1]{(\ref{#1})}

\begin{document}
%
\title{Event-based Dynamic Graph Representation Learning for Patent Application Trend Prediction}
%
%
%

\author{Tao~Zou, Le~Yu, Leilei~Sun, Bowen~Du, Deqing~Wang, Fuzhen~Zhuang
\IEEEcompsocitemizethanks{\IEEEcompsocthanksitem T. Zou, L. Yu, L. Sun, B. Du and D. Wang are with the SKLSDE and BDBC Lab, Beihang University, Beijing, 100191, China.\protect\\
E-mail: zoutao@buaa.edu.cn, yule@buaa.edu.cn, leileisun@buaa.edu.cn, dubowen@buaa.edu.cn, dqwang@buaa.edu.cn
\IEEEcompsocthanksitem F. Zhuang 
Institute of Artificial Intelligence, Beihang University, Beijing, 100191, China and SKLSDE, School of Computer Science, Beihang University, Beijing 100191, China.\protect\\
E-mail: zhuangfuzhen@buaa.edu.cn}

\thanks{(Corresponding author: Leilei Sun.)}}

\markboth{IEEE TRANSACTIONS ON KNOWLEDGE AND DATA ENGINEERING,~Vol.~XX, No.~X, XX~XXXX}%
{Yu \MakeLowercase{\textit{et al.}}:Event-based Dynamic Graph Representation Learning for Patent Application Trend Prediction}
%

\IEEEtitleabstractindextext{%
\begin{abstract}
Accurate prediction of what types of patents that companies will apply for in the next period of time can figure out their development strategies and help them discover potential partners or competitors in advance. 
Although important, this problem has been rarely studied in previous research due to the challenges in modeling companies’ continuously evolving preferences and capturing the semantic correlations of classification codes. 
To fill this gap, we propose an event-based dynamic graph learning framework for patent application trend prediction. In particular, our method is founded on the memorable representations of both companies and patent classification codes. When a new patent is observed, the representations of the related companies and classification codes are updated according to the historical memories and the currently encoded messages. Moreover, a hierarchical message passing mechanism is provided to capture the semantic proximities of patent classification codes by updating their representations along the hierarchical taxonomy. Finally, the patent application trend is predicted by aggregating the representations of the target company and classification codes from static, dynamic and hierarchical perspectives. Experiments on real-world data demonstrate the effectiveness of our approach under various experimental conditions, and also reveal the abilities of our method in learning semantics of classification codes and tracking technology developing trajectories of companies.
\end{abstract}

\begin{IEEEkeywords}
Patent application trend, classification codes, dynamic representations, hierarchical taxonomy
\end{IEEEkeywords}}

\maketitle

\IEEEdisplaynontitleabstractindextext

%
\IEEEpeerreviewmaketitle

\section{Introduction}
\label{section-1}
\IEEEPARstart{A}{} patent is an exclusive right granted for an invention that is never published anywhere else. Generally, a patent can be applied by single or multiple owners and is assigned with one or multiple classification codes, which are hierarchically organized as a taxonomy system. Classification codes in the adjacent levels are connected with each other and have similar concepts \cite{DBLP:books/sp/11/HarrisAS11}. According to the report from World Intellectual Property Organization (WIPO)\footnote{\url{https://www.wipo.int/edocs/pubdocs/en/wipo_pub_943_2021.pdf}}, the number of patent applications is increasing rapidly and has reached about 3.28 million in 2020, which demonstrates the growing importance of patents. 

\begin{figure}[!htbp]
    \centering
    \includegraphics[width=1.00\columnwidth]{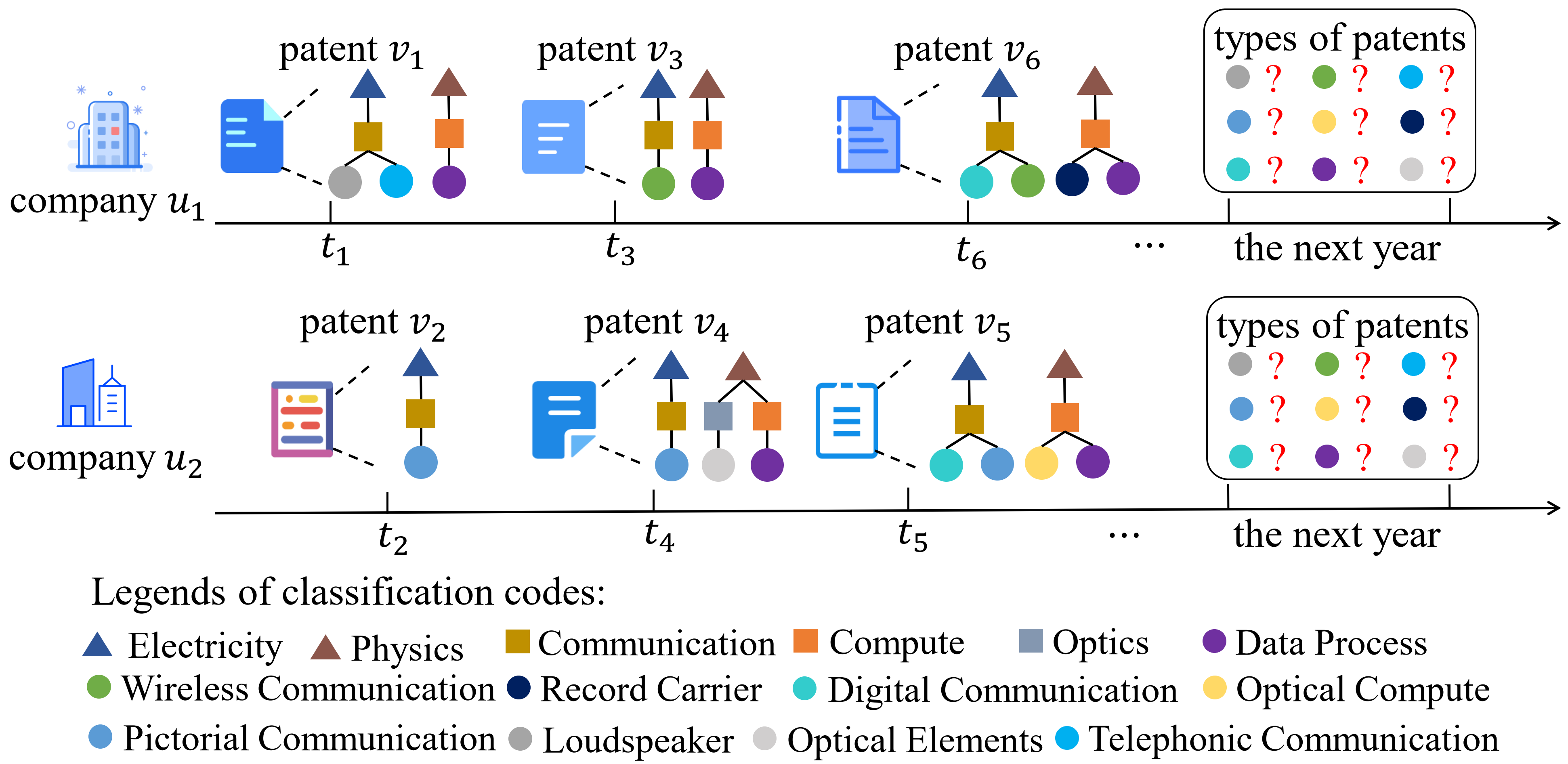}
    \caption{Given the sequences of patents applied by company $u_1$ and company $u_2$, the task of patent application trend prediction aims to predict what types (e.g. fields in technology) of patents that $u_1$ and $u_2$ will apply for in the next period of time. 
    \label{fig:motivation}
    }
\end{figure}

For companies, applying for patents is one of the key solutions for protecting their intellectual properties. Therefore, patents are the appropriate sources to analyze companies. Accurately predicting the patent application trends could help companies figure out development strategies and discover their potential partners or competitors in advance. As shown in \figref{fig:motivation}, we can observe that company $u_2$ shows a developing trend in the ``Physics - Compute - Data Process'' fields, which are the key research directions of the company $u_1$. If one could make the accurate prediction of patent application trends for companies, $u_1$ is able to make strategies against $u_2$ in advance. Besides, classification codes are organized as a taxonomy system, which helps us capture the semantic structural information among classification codes. For example, the “Digital Communication” field, “Telephonic Communication” field, and “Pictorial Communication” field are closer in semantic proximity since they belong to “Communication” field than the “Optical Compute” field and “Digital Communication” field.

In recent years, some approaches have been proposed for the predictive modeling of patents, which mainly focused on the following two tasks. Methods for the first task leveraged patents to predict emerging technologies. For example, \cite{DBLP:journals/corr/abs-1206-3933}, \cite{yoon2018exploring} predicted the emerging new technologies by analyzing the evolution of clusters in patents or visualizing patent information in the patent citation network. \cite{yoon2021doc2vec} encoded the patent information with Doc2vec algorithm to predict the future development direction of technology in the patent network. Methods for the second task aimed to predict future citations for patents. \cite{DBLP:conf/aaai/LiuYXWZC17} predicted the patent citations with a point process based on the patent citation sequences and textual information of patents. \cite{DBLP:conf/ijcai/JiCSFLR19} designed an attention-based model to capture the dynamic trends in patent citation sequences. However, these studies were designed for modeling in patents, and therefore cannot be applied to analyzing companies.

In this paper, we present a new task: predicting the patent application trend for a given company based on its historical application records, which is achieved by predicting the set of assigned classification codes of patents that the company will apply for in the future. We can infer the patent application trends of companies by predicting classification codes because classification codes stand for the high-level abstractions of patents and can well represent the technical directions of companies. Therefore, there are strong connections between company analysis and classification codes prediction. However, such a task is rather challenging due to two reasons. Firstly, the preferences of companies' technology development are continuously evolving over time with complicated temporal patterns. \textit{How to mine the latent evolving preferences from the previously applied patents is nontrivial}. {Secondly, classification codes are organized in a semantic taxonomy hierarchically, which indicates that some classification codes may be semantically close to each other. \textit{How to capture the semantic correlations of classification codes is another difficulty}.}

To tackle the above issues, we propose an \textbf{E}vent-based \textbf{D}ynamic \textbf{G}raph learning framework for \textbf{P}atent \textbf{A}pplication \textbf{T}rend prediction (EDGPAT). Our approach is developed to maintain the memorable representations of both companies and patent classification codes. 
When a new patent is observed, we first compute the messages of the related companies and classification codes via message encoders, and then dynamically update their representations based on the historical memories and the encoded messages. Moreover, we devise a hierarchical message passing mechanism to capture the semantic proximities of classification codes, which allows the classification codes in different levels to interact with each other along the hierarchical taxonomy. Finally, we present a multi-perspective representations fusing module to predict the patent application trend via aggregating representations of the target company and classification codes from static, dynamic, and hierarchical perspectives. Extensive experiments on real-world datasets show the superiority of our approach under a variety of experimental settings. The potential of our method in capturing the semantics of classification codes and tracking the technology-developing trajectories of companies is also demonstrated. 
Our key contributions include:

\begin{itemize}
    \item 
    {An event-based continuous-time representation learning framework is proposed for the new task, i.e., patent application trend prediction. Compared to classical user behavior modeling methods, we maintain memorable representations for companies and classification codes and apply a continuous-time updating mechanism, which could not only learn the companies' evolving preferences on technology development but also capture the semantic shifting of classification codes.}
            
    \item 
    {A hierarchical message passing mechanism is provided to learn the semantic correlations across classification codes. Different from the existing works on sequence behavior modeling, we utilize the hierarchical taxonomy structure and propagate their memorable representations along the hierarchical taxonomy.}
    
    \item 
    {A multi-perspective representations fusing component is designed to adaptively aggregate the representations of the target company and classification codes from the static, dynamic as well as hierarchical perspectives for predicting the patent application trend.}
\end{itemize}

\section{Related work}
\label{section-5}
In this section, we review the related research and discuss the differences between previous studies with our work.

\textbf{Patent Data Mining}. 
Patent documents provide information about technological developments and potential innovation trends in society. Over the past decades, a great number of efforts have been made on mining patents with the aim to discover their potentially great value \cite{DBLP:journals/sigkdd/ZhangLL14, krestel2021survey}. 
One branch of the existing methods mainly focused on the predictive modeling of patents, including the tasks of patent-based new technology prediction \cite{DBLP:journals/corr/abs-1206-3933,baumann2021comparative,yoon2021doc2vec}, and patent citation prediction \cite{DBLP:conf/aaai/LiuYXWZC17,DBLP:conf/ijcai/JiCSFLR19}. In particular, approaches for the first task predicted the emerging technologies according to the patent network \cite{DBLP:journals/corr/abs-1206-3933,yoon2018exploring} or the textual information of patents \cite{yoon2021doc2vec}. Methods for the second task captured temporal dependencies from the patent citation sequences by a point process \cite{DBLP:conf/aaai/LiuYXWZC17} or an attention mechanism \cite{DBLP:conf/ijcai/JiCSFLR19}. Another branch aimed to make patents analysis more automatic, such as patent classification \cite{DBLP:conf/bigcomp/RoudsariALL20, DBLP:conf/aaai/Tang0XPWC20}, patent retrieval \cite{DBLP:journals/kais/ShalabyZ19}, and patent text generation \cite{DBLP:journals/corr/abs-1907-02052}. Specifically, these tasks mostly focused on capturing semantic relationships from patent descriptions with text representation learning methods in Natural Language Processing (NLP) field. For example, \cite{DBLP:conf/bigcomp/RoudsariALL20} utilized Bert \cite{DBLP:conf/naacl/DevlinCLT19} pre-trained model to capture semantic dependencies in patent documents while \cite{DBLP:conf/aaai/Tang0XPWC20} combined GCN and attention mechanism to embed patent representations for classification.

In this paper, we present a new problem that predicts patent application for companies, which is orthogonal to the existing studies on patent mining and could help companies make better decisions on selecting partners and discovering competitors. In practice, patent application prediction is more challenging due to the considerations of 1) the evolving temporal patterns of companies and the classification codes that patents are associated with; and 2) the tree-like structure of classification codes in a hierarchical taxonomy.

\textbf{Sequential Behavior Modelling}.
The purpose of sequential behavior modeling is to learn meaningful latent representations for the targets to reflect their sequential patterns \cite{DBLP:journals/tois/FangZSG20}. In recent years, lots of studies have been proposed to capture sequential behaviors in various scenarios. For example, in the field of recommendation systems, \cite{DBLP:conf/www/RendleFS10} and \cite{DBLP:journals/corr/HidasiKBT15} investigated the sequential behaviors of users by combining the matrix factorization and Markov chain or utilizing the RNN architecture. \cite{DBLP:conf/kdd/HuH19} and \cite{DBLP:conf/www/YuWS0L22} studied the dynamics of user behaviors in sequential sets. To capture the semantic information in sequences, \cite{DBLP:conf/cikm/BianZZCHYW21} applied the contrastive curriculum learning algorithm in modeling users' behaviors.
In the field of intelligent treatment, patient sequential behaviors were analyzed by supervised reinforcement learning \cite{DBLP:conf/kdd/WangZHZ18} and heterogeneous long short-term memory networks \cite{DBLP:conf/kdd/0001YSLQT18}. In the field of the social network, understanding and analyzing the future activities of users is beneficial for many applications such as advertising systems \cite{DBLP:journals/tist/WangYGYMLX22}, cascade prediction \cite{DBLP:conf/aaai/TangLHXZS21}. For example, \cite{DBLP:conf/cikm/LiGBC17} analyzed the correlations about online activities with Hawkes Process and \cite{DBLP:conf/kdd/LuoZYBYLQY20} combined heterogeneous graph neural network to predict the real-time customer response.

{In the studied problem, applying for patents can be treated as the sequential behaviors of each company. However, the existing sequential behavior modeling approaches are always capturing users' preference patterns from every single sequence. Compared to these works, we organize the sequences of all the companies as a universal sequence of chronological events and maintain memorable representations for companies and classification codes to learn the dynamic evolving preferences. Besides, we capture the hierarchical structure information among classification codes along the taxonomy system.}
 
\textbf{Graph Representation Learning}.
GNNs have shown superiority in graph representation learning and have been used in practical applications such as 
social network analysis \cite{DBLP:conf/kdd/QiuTMDW018,DBLP:conf/www/Fan0LHZTY19,DBLP:conf/pkdd/LuXSFWZL20}, and recommendation systems \cite{DBLP:conf/kdd/YingHCEHL18,DBLP:conf/sigir/Wang0WFC19,DBLP:conf/sigir/0001DWLZ020}. For example, \cite{DBLP:conf/sigir/Wang0WFC19} proposed NGCF, which captured the interactions between users and items with message propagation in the graph structure while \cite{DBLP:conf/sigir/0001DWLZ020} analyzed the GCN framework and designed a simplified GCN model for prediction. Besides, some work \cite{DBLP:journals/tnn/LiNCYZS18,DBLP:journals/tcyb/LuoCNYHZ18,DBLP:journals/tnn/ZhouCSSYN20} learned adaptive structure in graph representation learning. Since most of the existing GNNs are designed for static graphs, some recent approaches have extended GNNs to dynamic graphs to capture the dynamics in real-world scenarios \cite{DBLP:journals/jmlr/KazemiGJKSFP20,DBLP:journals/corr/abs-2303-13047}. One part of the methods learned on dynamic graphs based on a sequence of snapshots, where each snapshot corresponded to a static graph that contains the observations up to a certain time \cite{DBLP:conf/aaai/ParejaDCMSKKSL20,DBLP:conf/wsdm/SankarWGZY20,DBLP:conf/kdd/YuSDL0L20}. For example,  \cite{DBLP:conf/aaai/ParejaDCMSKKSL20} learned the evolving parameters for GCN in different timestamps with the LSTM network. Another part of the methods treated the dynamic process as an entire graph with fine-grained temporal information \cite{DBLP:conf/iclr/XuRKKA20,DBLP:conf/cikm/ChangLW0FS020} or a sequence of chronological events \cite{DBLP:journals/corr/abs-2006-10637,DBLP:conf/log/LuoL22}, which captures the continuous evolving characteristics over a period of time. 

The existing works are proposed for learning general dynamic representations, which are not specifically designed for patent application trend prediction. In this paper, we propose a customized continuous-time learning framework for patent application trend prediction.

\section{Preliminaries}
\label{section-2}
In this section, we first present some necessary definitions and then formalize the studied problem.

\subsection{Definitions}
Let $\mathbb{U}=\{u_1,u_2,\cdots,u_m\}$ and $\mathbb{V}=\{v_1,v_2,\cdots,v_n\}$ represent the collections of $m$ companies and $n$ lowest-level classification codes, respectively. Let $\mathbb{S}=\{s_{1,1},\cdots,s_{1,L},\cdots,s_{j,l}\cdots,s_{n,1},\cdots,s_{n,L}\}$ denote the collections of classification codes in all the levels, where classification code $s_{j,L}$ can reach $s_{j,l}$ via exactly $L-l$ hops from bottom to top in the hierarchical taxonomy. 
$L$ is the maximal depth. In fact, the lowest-level classification code $v_j \in \mathbb{V}$ is actually identical to the $L$-level $s_{j,L} \in \mathbb{S}$.

\begin{definition}
     \textbf{Event}.
     An event $e_k=(U_k,V_k,t_k)$ is defined as the application of patent $p_k$, where $U_k \subset \mathbb{U}$ is the set of companies that apply for patent $p_k$, $V_k\subset \mathbb{V}$ denotes the set of lowest-level classification codes that $p_k$ involves and $t_k \in \mathbb{R}^+$ is the applied timestamp of $p_k$.
\end{definition}

\begin{definition}
     \textbf{Event-based Dynamic Graph}. In this problem, we define an event-based dynamic graph as a set of chronological events $\mathcal{G}^t=\left\{e_1,\cdots,e_k\right\}$, where $0 \textless t_1 \leq \cdots \leq t_k \leq t$.  
     
     In our work, each event $e_i=(U_i,V_i,t_i)$ could be treated as an interaction between companies $U_i$ and classification codes $V_i$. Hence, companies $U_i$ and classification codes $V_i$ are defined as the nodes and the interactions between $U_i$ and $V_i$ represent edges in our dynamic graph.
\end{definition}

\subsection{Problem Formalization}
\textbf{Patent Application Trend Prediction} aims to learn a function $f\left(\cdot\right)$ to predict the \textit{lowest-level classification codes} (which reflect the most fine-grained information) of future patents that will be applied by any company $u_i \in \mathbb{U}$ in the next period of time, which is set to the next year in this paper. We solve such a problem by learning on the event-based dynamic graph and hierarchical taxonomy via
$$\hat{p}_{i,j}=f\left(\mathcal{G}^{T_i}, \mathbb{S}\right),$$
where $\hat{p}_{i,j}$ denotes the probability that company $u_i$ will apply patents with the lowest-level classification code $v_j \in \mathbb{V}$. $T_i$ is the applied timestamp of company $u_i$'s last patent.

\section{Methodology}
\label{section-3}

\begin{figure*}[!htbp]
    \centering
    \includegraphics[width=2.0\columnwidth]{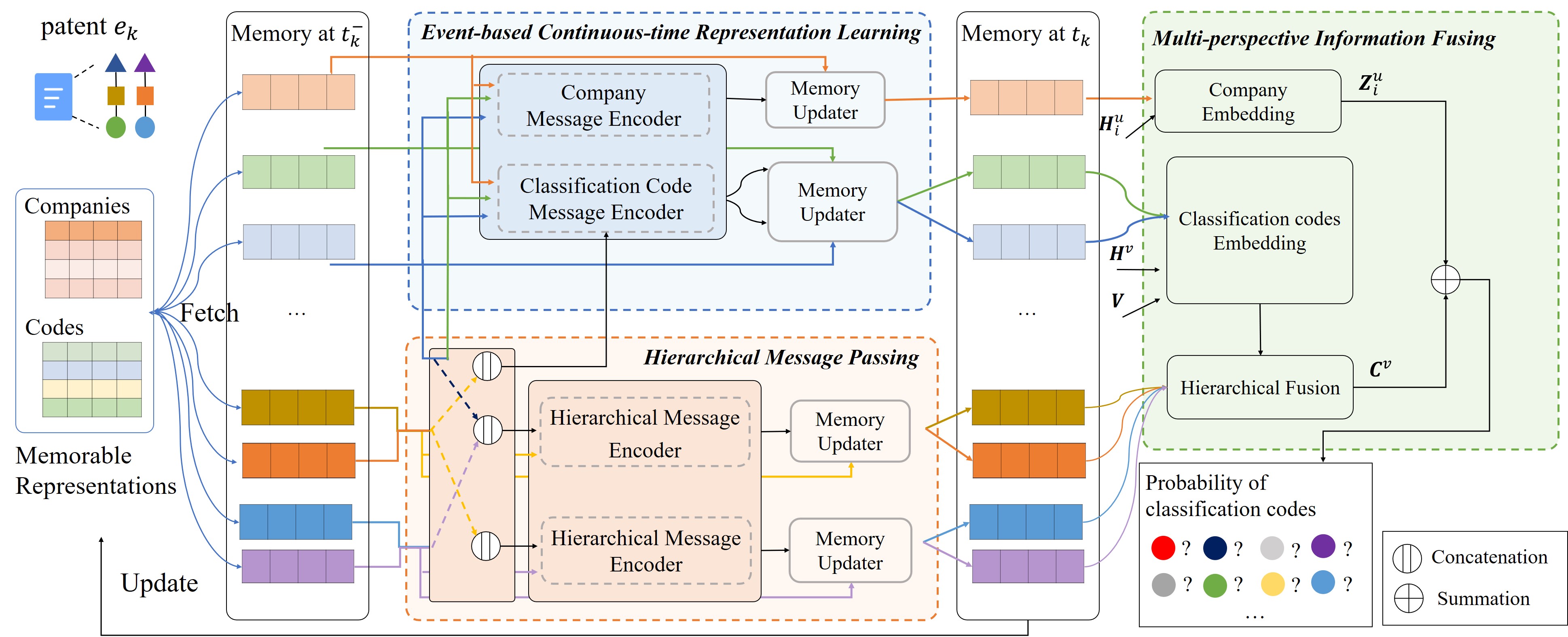}
    \caption{Framework of the proposed model. We just show the calculations of the patent classification codes and one of the related companies for simplicity.}
    \label{fig:framework}
\end{figure*}

In this section, we first introduce the framework of our proposed model and then present each component step by step. As shown in \figref{fig:framework}, our model is composed of three components: event-based continuous-time representation learning, hierarchical message passing, and multi-perspective representations fusing, which aims to maintain memorable representations of both companies and patent classification codes.
Specifically, when a new patent is observed, the first component encodes the messages of the related companies and patent classification codes, and dynamically updates their memorable representations based on the previous memories and currently encoded messages. The second component captures the semantic correlations of classification codes by allowing classification codes in the adjacent levels to interact with each other along the hierarchical taxonomy. The last part computes the appearing probabilities of the lowest-level classification codes for patent application trend prediction, which simultaneously aggregates the representations of the target company and classification codes from static, dynamic, and hierarchical perspectives.

\subsection{Event-based Continuous-time Representation Learning}
\label{section-3-1}
A unique perspective of our approach is to represent each patent application as an event and learn the dynamic memorable representations of both companies and patent classification codes from the event-based dynamic graph, which corresponds to the sequence of chronological events. 

As we aim to predict the patent application trend for each company, when a new patent is observed, we first divide it into several items, where each item records the information of an individual company and  the set of related classification codes. For example, a new patent application event $e_k=\left(U_k,V_k,t_k\right)$ with $U_k=\{u_1,u_2,u_3\}$ will be split into the following three items: $\left(u_1,V_k,t_k\right)$, $\left(u_2,V_k,t_k\right)$, $\left(u_3,V_k,t_k\right)$. Then, our approach learns from the three items and makes predictions for $u_1$, $u_2$, and $u_3$, respectively. We use $\bm{u}_{i}^{t},\bm{v}_{j}^{t} \in \mathbb{R}^{d}$ to denote the memorable representation of company $u_i$ and classification code $v_j$ at timestamp $t$, which records their historical states. When the patent application event $e_{k}$ is observed, the latest memorable representations of company $u_i\in U_{k}$ and the classification codes in $V_{k}$ will be updated via the message encoders and memory updaters. 

\subsubsection{Representation Learning for Companies}
When a patent $e_{k}$ is applied, we first compute the message $\bm{m}^{u}_{i,k}$ for company $u_i$. Then, we update the memorable representation with $\bm{m}^{u}_{i,k}$ and historical memory $\bm{u}_{i}^{t_{k}^{-}(u_i)}$, where ${t_{k}^{-}(u_i)}$ is the latest time that company $u_i$ applied a patent right before $t_k$. 

\textbf{Company Message Encoder}. Formally, $\bm{m}^{u}_{i,k}$ is computed by three parts: the integrated information $\bm{h}_{V_k}$ of classification codes in $V_{k}$, the historical memory $\bm{u}_{i}^{t_{k}^{-}(u_i)}$, and the temporal encoding $\bm{h}_{{\Delta{t}_k}(u_i)}$. 
Specifically, $\bm{h}_{V_k}$ contains the information of all the classification codes in $V_{k}$, which is calculated by
\begin{equation}
\label{equ:information for classification codes}
    \bm{h}_{V_k} = g\left(\bm{v}_{c_1}^{t_k^-(v_{c_1})}\|\cdots\|\bm{v}_{c_k}^{t_k^-(v_{c_k})}\right),
\end{equation}
where $\bm{v}_{c_j}^{t_k^-(v_{c_j})}$ denotes the latest memories of the lowest-level classification code $v_{c_j} \in V_k$ before time $t_{k}$. $g(\cdot)$ denotes the information integrating encoder, which is implemented by the mean pooling operation in this work. 

When predicting the patent application trend, timestamps information plays an essential role in reflecting the temporal patterns. Inspired by \cite{DBLP:conf/iclr/XuRKKA20}, we design a trainable time-encoding function with $\cos(\cdot)$ and $\sin(\cdot)$ functions, which maps the time interval $\Delta{t_k}(u_i)=t_k - t_k^-(u_i)$ into a continuous vector and learns the temporal patterns by 

\begin{small}
\begin{equation}
\label{equ:temporal_representation}
\begin{split}
   &\bm{h}_{\Delta{t_k}(u_i)} =\sqrt{\frac{1}{d_T/2}}\bigg[\cos\left(w_1 \Delta{t_k}(u_i)\!+\!b_1\right),\sin\left(w_1 \Delta{t_k}(u_i) \!+\!b_1\right),\\
   &\cdots,\cos\left(w_{d_T/2} \Delta{t_k}(u_i) \!+\!b_{d_T/2} \right),\sin\left(w_{d_T/2} \Delta{t_k}(u_i) \!+\!b_{d_T/2} \right)\bigg],
\end{split}
\end{equation}
\end{small}
where $\bm{w}=[w_1,\cdots,w_{d_T/2}]$ and $\bm{b}=[b_1,\cdots,b_{d_T/2}]$ are learnable parameters. $\bm{h}_{\Delta{t_k}(u_i)}$ is a $d_T$-dimensional vector and $d_{T}$ is the dimension of temporal encoding, which we set to $d$ in this paper.

Finally, we concatenate $\bm{h}_{V_k}$, $\bm{u}_{i}^{t_{k}^{-}(u_i)}$ and $\bm{h}_{{\Delta{t}_k}(u_i)}$ to generate the encoded message for company $u_i$ at $t_k$ via

\begin{equation}
\label{equ:company message}
     \bm{m}_{i,k}^{u} = {MSG_u}\left(\bm{h}_{V_k} \| \bm{u}_{i}^{t_{k}^{-}(u_i)} \|  \bm{h}_{\Delta{t_k(u_i)}}\right),
\end{equation}
where ${MSG_u}(\cdot)$ is the message encoder for companies, and we implement it by a 2-layer perceptron neural network.

\textbf{Company Memory Updater}. After obtaining the encoded message $\bm{m}_{i,k}^{u}$ of company $u_i$, we design a memory updater to keep the memorable representation of $u_i$ up to date, which is realized by combining $\bm{m}_{i,k}^{u}$ and $u_i$'s historical memory $\bm{u}_{i}^{t_{k}^{-}(u_i)}$ with the Gated Recurrent Unit (GRU) \cite{DBLP:conf/emnlp/ChoMGBBSB14},
\begin{equation}
\label{equ:company memory update}
    \bm{u}_{i}^{t_{k}} = GRU_{u}\left(\bm{m}_{i,k}^{u},\bm{u}_{i}^{t_{k}^{-}(u_i)}\right).
\end{equation}

\subsubsection{Representation Learning for Lowest-level Classification Codes} Similar to the representation learning for companies, we also learn the memorable representations of the lowest-level classification codes in $V_{k}$. For classification code $v_j$ in $V_{k}$, we first encode its message $\bm{m}_{j,k}^{v}$. Then, we update the memorable representation with $\bm{m}_{j,k}^{v}$ and historical memory $\bm{v}_{j}^{t_{k}^{-}(v_j)}$, where $t_{k}^{-}(v_j)$ is the latest time that a patent was applied right before timestamp $t_k$ and assigned with classification code $v_j$.

\textbf{Lowest-level Classification Code Message Encoder}.
For classification code $v_j$, message $\bm{m}_{j,k}^{v}$ is computed by four parts: the information from the related company $u_i$, the historical memory $\bm{v}_{j}^{t_{k}^{-}(v_j)}$, temporal encoding $\bm{h}_{\Delta{t_k(v_j)}}$ and hierarchical information $\bm{h}_{j,L,k}$. In particular, we utilize $\bm{u}_{i}^{t_{k}^{-}(u_i)}$ as the memorable representation of the interacted company $u_i$. For temporal encoding $\bm{h}_{\Delta{t_k(v_j)}}$, we use $\Delta{t_{k}}(v_j) = t_k - t_{k}^{-}(v_j)$ as the time interval and learn it by the encoding function in \equref{equ:temporal_representation}. We derive the hierarchical information $\bm{h}_{j,L,k}$ along the hierarchical taxonomy, which will be introduced in \secref{subsec:hierarchy}. Finally, we concatenate the obtained $\bm{u}_{i}^{t_{k}^{-}(u_i)}$, $\bm{v}_{j}^{t_{k}^{-}(v_j)}$, $\bm{h}_{\Delta{t_k(v_j)}}$ and $\bm{h}_{j,L,k}$, and encode message for classification code $v_j$ at timestamp $t_k$ as follows,
\begin{equation}
\label{equ:classification codes message}
    \bm{m}_{j,k}^{v} = MSG_v\left(\bm{u}_{i}^{t_{k}^{-}(u_i)} \| \bm{v}_{j}^{t_{k}^{-}(v_j)} \| \bm{h}_{\Delta{t_k(v_j)}} \| \bm{h}_{j,L,k}\right), 
\end{equation}
where $MSG_v(\cdot)$ is the message encoder for lowest-level classification codes, which is also implemented by a 2-layer perceptron neural network.

\textbf{Lowest-level Classification Code Memory Updater}. After obtaining the encoded message $\bm{m}_{j,k}^{v}$ of classification code $v_j$, we design a memory updater to keep the memory of $v_j$ up to date, which is calculated by the message $\bm{m}_{j,k}^{v}$ and its historical memory $\bm{v}_{j}^{t_{k}^{-}(v_j)}$ with a GRU model,
\begin{equation}
\label{equ:classification codes update}
    \bm{v}_{j}^{t_{k}} = GRU_{v}\left(\bm{m}_{j,k}^{v},\bm{v}_{j}^{t_{k}^{-}(v_j)}\right).
\end{equation}

\subsection{Hierarchical Message Passing}
\label{subsec:hierarchy}
As stated in \secref{section-1}, classification codes are organized as a hierarchical taxonomy, in which every classification code is coherent with its ancestor and two classification codes are semantically similar if they belong to the same ancestor. To this end, we devise a hierarchical message passing mechanism to automatically capture the semantic correlations of classification codes by propagating memorable representations along the hierarchical taxonomy. Similar to the above calculating process, higher-level classification codes in $\mathbb{S}$ also have memorable representations and are computed by the message aggregating and memory updating process.

\textbf{Higher-level Classification Code Message Encoder}.
We define $\bm{s}_{j,l}^t$ as the memorable representation of the $l$-level classification code $s_{j,l}$ at timestamp t, where $s_{j,L}$ can reach the $l$-level classification code  $s_{j,l}$ via exactly $L-l$ hops, and $1 \leq l \leq L$. We encode message $\bm{h}_{j,l,k}$ of $s_{j,l}$ at timestamp $t_{k}$ along the hierarchical taxonomy. We first combine the memorable representations of classification codes at the $(l-1)$-th and $(l+1)$-th levels, that is,
\begin{equation}
\label{equ:hierarchical information}
	\bm{h}_{j,l,k}  = 
		\bm{s}_{j,l-1}^{t_{k}^{-}(s_{j,l-1})} \| \bm{s}_{j,l+1}^{t_{k}^{-}(s_{j,l+1})}.
\end{equation}
It is worth noticing that classification codes at the highest $1$-th level only receive memories from the $2$-nd level, and classification codes at the lowest $L$-th level only obtain memories at the $(L-1)$-th level (as stated in \secref{section-3-1}). 

Then, the representation of the time interval of $s_{j,l}$ is identically calculated via \equref{equ:temporal_representation}, and we denote it by $\bm{h}_{\Delta{t_k}(s_{j,l})}$. Finally, we concatenate $\bm{h}_{j,l,k}$, $\bm{s}_{j,l}^{t_{k}^{-}(s_{j,l})}$ and $\bm{h}_{\Delta{t_k}(s_{j,l})}$, and encode the message for $s_{j,l}$ ($1\leq l<L$) at timestamp $t_k$ by
\begin{equation}
\label{equ:higher classification codes message}
     \bm{m}_{j,l,k}^{s} = {MSG_{s_{l}}}\left(\bm{h}_{j,l,k} \| \bm{s}_{j,l}^{t_{k}^{-}(s_{j,l})} \|  \bm{h}_{\Delta{t_k}(s_{j,l})}\right), 
\end{equation}
where $MSG_{s_{l}}$ is the message encoder for classification codes in the $l$-level, corresponding to a 2-layer perceptron neural network.

\textbf{Higher-level Classification Code Memory Updater}. The memorable representation of $s_{j,l}$ is updated to be the latest via another GRU model by,
\begin{equation}
\label{equ:higher classification codes updater}
     \bm{s}_{j,l}^{t_{k}} = GRU_{s_l}\left(\bm{m}_{j,l, k}^{s},\bm{s}_{j,l}^{t_{k}^{-}(s_{j,l})}\right).
\end{equation}

\subsection{Multi-perspective Representations Fusing}
We compute the probability of all the lowest-level classification codes that will be assigned in the future patents for company $u_i$ by aggregating the representations of company $u_i$ and classification codes from dynamic, static, and hierarchical perspectives. The static information for companies and lowest-level classification codes are denoted as $\bm{H}^{u},\bm{H}^{v}$, which are randomly initialized and can be optimized in the training process. 

Firstly, we compute the preferences for company $u_i$ with a weighted aggregation of its curent memory $\bm{u}_i^{t_k}$ and static information $\bm{H}^{u}_{i}$,
\begin{equation}
    \label{equ:company_preference}
    \bm{Z}_{i,k}^{u} = MLP(\alpha_{i}^{u} \bm{u}_i^{t_k} + (1-\alpha_{i}^{u}) \bm{H}^{u}_{i}),
\end{equation}
where $\bm{\alpha}^{u} \in \mathbb{R}^m$ is the learnable parameter for company embedding. $\bm{Z}_{i,k}^{u}\in \mathbb{R}$ denotes the preferences of company $u_i$ at timestamp $t_k$.

Secondly, the predicted representations of all the lowest-level classification codes are computed from static, dynamic, and hierarchical information. We denote the lowest-level classification codes involved in company $u_i$'s historical patents up to timestamp $t_k$ as $Q_{i,t_k}^{v}$. According to the sequence of company $u_i$' patents, lowest-level classification codes are divided into three parts: 1) the ones that are involved in the current patent $p_k$'s classification codes $V_k$; 2) the ones belonging to collection $Q_{i,t_k^-}^{v}\setminus V_{k}$ that appeared in the company $u_i$'s historical patents and not in patent $p_k$, namely historical classification codes; 3) the remaining ones in $V\setminus Q_{i,t_k}^{v}$ which have never been interacted with company $u_i$ up to timestamp $t_k$. 

To provide the final predictions, we design different strategies for classification codes in the aforementioned three parts. For classification codes in the first part, we combine the \textit{updated memories} and static information for their representations. For classification codes in the second part, we utilize a 2-layer perceptron network to embed their \textit{historical memories} and aggregate them with the related static information. For those in the third part, we use static information as their predicted representations. 
To sum up, the process for calculating representations of all classification codes is,
\begin{equation}
\begin{split}
\label{equ:classification embedding}
\bm{Z}^{v}_{j,k} &= 
    \left(1- \beta_{j} \alpha_{j}^{v}\right) \bm{H}^{v}_{j} \\ &+ \beta_{j} \alpha_{j}^{v} \left(\gamma_j \bm{v}_j^{t_k}  + \left(1- \gamma_j\right) MLP\left(\bm{v}_j^{t_k^-(v_j)}\right)\right),
\end{split}
\end{equation}
where $\bm{H}^{v}_{j}$ denotes static information for $v_j$, $\bm{\alpha}^{v} \in \mathbb{R}^{n}$ is the trainable parameter for lowest-level classification codes. $\bm{\beta} \in \mathbb{R}^{n}$ and $\bm{\gamma} \in \mathbb{R}^{n}$ are two indicator vectors to distinguish classification codes from the three parts.
In particular, $\beta_{j}$ is set to 1 when $v_j \in Q^{v}_{i,t_k}$, otherwise 0. $\gamma_{j}$ is set to 1 when $v_j \in V_k$, otherwise 0.

We further incorporate hierarchical information to compute the predicted representations of classification codes. We aggregate $\bm{Z}^{v}_{j,k}$ and the corresponding hierarchical memories, 
\begin{equation}
\label{equ:hierarchical embedding}
\begin{split}
    \bm{C}_{j,k}^{v} = \lambda_L \bm{Z}_{j,k}^{v} + \sum_{l=1}^{L-1} \lambda_l \bm{s}_{j,l}^{t_k},
\end{split}
\end{equation}
where $\bm{\lambda} \in \mathbb{R}^{L}$ is the trainable vector for each level's classification codes in the hierarchical taxonomy. 


Finally, we compute the probability of the lowest-level classification code $v_j$ for company $u_i$ based on its preference $\bm{Z}_{i,k}^{u}$ in \equref{equ:company_preference} and $v_j$'s predicted representation $\bm{C}_{j,k}^{v}$,
\begin{equation}
\label{equ:probability produce}
    \hat{p}_{i,j} = MLP\left(\bm{Z}_{i,k}^{u} + \bm{C}_{j,k}^{v}\right).
\end{equation}

\begin{algorithm}[!htbp]
\SetKwComment{Comment}{/* }{ */}
\SetKwInOut{Input}{Input}
\SetKwInOut{Output}{Output}
\caption{Training process of EDGPAT}
\label{alg:training_process}
\Input{Collection of companies $\mathbb{U}$ and \textit{lowest-level classification codes} $\mathbb{V}$, a sequence of chronological events $\mathcal{G}^t=\left\{e_1,\cdots,e_k\right\}$ with $0 < t_1 \leq \cdots \leq t_k \leq t$, maximum number of training epochs $MaxEpoch$\;}
\Output{The model parameters $\Theta$ after training\;}
Initialize the parameters in EDGPAT with random weights $\Theta$\ and  set $Epochs \gets 1$\;
Initialize the memories of all the companies and classification codes in hierarchical taxonomy with zero vectors\;
\While{not converge and $Epochs \leq MaxEpoch$}{
    \For{batch $B_b \in \mathcal{B}$ }{
        $\{\bm{m}_{i,k}^{u}\}$ $\gets$ Encode the messages of companies in $B_b$ via \equref{equ:information for classification codes}-\myref{equ:company message} with $\{\bm{u}_i^{t_k^-(u_i)}\}$ and $\{\bm{v}_j^{t_k^-(v_j)}\}$ as inputs\;
        
        $\{\bm{m}_{j,k}^{v}\}$ $\gets$ Encode the messages of \textit{lowest-level} classification codes in $B_b$ via \equref{equ:temporal_representation}, \myref{equ:classification codes message}, \myref{equ:hierarchical information} with $\{\bm{u}_i^{t_k^-(u_i)}\}$, $\{\bm{v}_j^{t_k^-(v_j)}\}$ and $\{\bm{s}_{j,L-1}^{t_k^-(s_{j,L-1})}\}$ as inputs\;
        
        $\{\bm{m}_{j,l,k}^{s}\}$ $\gets$ Encode the messages of \textit{higher-level} classification codes in $B_b$ via  \equref{equ:temporal_representation}, \myref{equ:hierarchical information}, \myref{equ:higher classification codes message} with $\{\bm{s}_{j,l-1}^{t_{k}^{-}(s_{j,l-1})}\}$, $\{\bm{s}_{j,l}^{t_{k}^{-}(s_{j,l})}\}$ and $\{\bm{s}_{j,l+1}^{t_{k}^{-}(s_{j,l+1})}\}$ as inputs\;
        
        $\{\bm{u}_i^{t_k}\}$ $\gets$ Update the memories of companies in $B_b$ via \equref{equ:company memory update} with $\{\bm{m}_{i,k}^{u}\}$ and $\{\bm{u}_i^{t_k^-(u_i)}\}$ as inputs\;
        
        $\{\bm{v}_j^{t_k}\}$ $\gets$ Update the memories of \textit{lowest-level} classification codes in $B_b$ via \equref{equ:classification codes update} with $\{\bm{m}_{j,k}^{v}\}$ and $\{\bm{v}_j^{t_k^-(v_j)}\}$ as inputs\;
        
        $\{\bm{s}_{j,l}^{t_{k}}\}$ $\gets$ Update the memories of \textit{higher-level} classification codes in $B_b$ via \equref{equ:higher classification codes updater} with $\{\bm{m}_{j,l,k}^{s}\}$ and $\{\bm{s}_{j,l}^{t_{k}^{-}(s_{j,l})}\}$ as inputs\;
        
        $\{\hat{p}_{i,j}\}$ $\gets$ Compute the probabilities by fusing dynamic, static embedding via \equref{equ:company_preference}-\myref{equ:probability produce} with $\{\bm{u}_i^{t_k}\}$, $\{\bm{v}_j^{t_k}\}$, $\{\bm{s}_{j,l}^{t_k}\}$, $\{\bm{H}^{u}_{i}\}$, $\{\bm{H}^{v}_{j}\}$, $\bm{\alpha}^u$, $\bm{\alpha}^v$ and $\bm{\lambda}$ as inputs\;
        Optimize the model parameters $\Theta$ by backpropagation via \equref{equ:optimization}\;
    }
    $Epochs \gets Epochs + 1$\;
}
\end{algorithm}

\subsection{Model Training Process}
We initialize the memorable representations of all the companies and classification codes in this work as zero vectors. We treat the task of patent application trend prediction as a multi-label classification problem, where each lowest-level classification code represents a label. The ground truth of company $u_i$ is defined as $\bm{p}_{i}\in\{0,1\}^n$, where the value of $1$ for $p_{i,j}$ represents that classification code $v_j$ appears in future patents of the company $u_i$. To train our model, the multi-label classification problem is further converted into a multiple binary classification problem and the model can be optimized by minimizing the cross-entropy loss,
\begin{equation}
\label{equ:optimization}
    L = -\sum_{u_i \in \mathbb{U}}\sum_{v_j \in \mathbb{V}} p_{i,j}\log(\hat{p}_{i,j}) + (1-p_{i,j})\log(1-\hat{p}_{i,j}).
\end{equation}
\algoref{alg:training_process} shows the training process of our model.

\section{Experiments}
\label{section-4}
In this section, we conduct extensive experiments on real-world datasets for model evaluation. 
    

        

\subsection{Descriptions of Datasets}
We use the patent application records of technology companies in China from 2010 to 2018 to conduct experiments. The patents cover a variety of research fields, including physics, mechanism, medicine, biology, chemistry, etc. For each patent, we extract its related companies and classification codes. {We obtain the hierarchical taxonomy from the International Patent Classification (IPC)\footnote{\url{https://ipcpub.wipo.int}}, which organizes classification codes into five levels.} We create the following four datasets with the increasing number of research fields. \textbf{PHYSICS} and \textbf{MECHANISM} contain patents that respectively belong to the physics and mechanism fields since these two fields have the largest numbers of lowest-level classification codes in all the fields. \textbf{P \& M} contains patents that from physics or mechanism fields. \textbf{ALL} contains patents of all the research fields, including ``Necessities", ``Transportation", ``Chemistry", ``Textile", ``Construction", ``Mechanism", ``Physics" and ``Electricity". From \tabref{tab:datasets_information}, we could observe that the maximum number and minimum of classification codes in L5 level in each field are 2283 and 15811 in \textbf{ALL}, which illustrates the different structure in each field and the importance of capturing semantic information among classification codes.

For data partition, we use patent application records in 2010-2016, 2017, and 2018 for training, validation, and testing, respectively. We drop the companies that have no historical records before 2016 or have no patent application records in 2016, 2017, and 2018. Statistics of the datasets are in \tabref{tab:datasets_information}, where \#com represents the number of companies. L1, L2, L3, L4 and L5 denote the 1st-level, 2nd-level, 3rd-level, 4-th level, 5th-level.

\begin{table}[!htbp]
\centering
\caption{Statistics of the datasets.}
\label{tab:datasets_information}
\resizebox{\columnwidth}{!}
{
\setlength{\tabcolsep}{1.2mm}
{

\begin{tabular}{c|c|c|ccccc}
\hline
\multirow{2}{*}{Datasets} & \multirow{2}{*}{\#com} & \multirow{2}{*}{\#patents} & \multicolumn{5}{c}{\#classification codes} \\ \cline{4-8} 
                          &                        &                            & L5       & L4       & L3      & L2   & L1  \\ \hline
PHYSICS                   & 2,445                  & 234,241                    & 6,146    & 932      & 146     & 14   & 1   \\
MECHANISM                 & 1,769                  & 167,203                    & 6,266    & 1,361    & 186     & 17   & 1   \\
P \& M                    & 12,952                 & 396,888                    & 12,952   & 2,360    & 333     & 31   & 2   \\
ALL                       & 14,695                 & 1,263,600                  & 60,082   & 10,049   & 1,204   & 124  & 8   \\
\hline
\end{tabular}

}
}
\end{table}

\subsection{Compared Methods}
We compare our approach with statistical methods (TOP and PersonalTOP), classical machine learning methods (SVM and RF), and deep learning methods (FPMC, DREAM, GRU4Rec, TGAT, and TGN).
\begin{itemize}
	\item  \textbf{TOP} collects the most frequent classification codes in all the patents as the prediction results for any company.

    \item  \textbf{PersonalTOP} uses the most frequent classification codes in the personalized sequence for each company.

    \item  \textbf{SVM} constructs the hyperplanes in a high-dimensional space for classification code prediction \cite{DBLP:books/cu/CS2000}.

    \item  \textbf{RF} builds multiple decision trees and outputs the classification codes selected by the majority of trees \cite{DBLP:journals/ml/Breiman01}.
    

    \item  \textbf{FPMC} proposes a combination of matrix factorization and Markov chain to capture the sequential dependencies and long-term user preferences \cite{DBLP:conf/www/RendleFS10}.
    
    \item  \textbf{DREAM} obtains user representations by the pooling operation over the interacted items during a specific period and feeds the dynamic representations into the Recurrent Neural Network (RNN) architecture to make predictions \cite{DBLP:conf/sigir/YuLWWT16}.
    
    \item \textbf{GRU4Rec} encodes the items with an embedding module and predicts the next-period items with the Gated Recurrent Unit (GRU) neural networks \cite{DBLP:journals/corr/HidasiKBT15}.

    \item  \textbf{TGAT} considers the temporal dependencies when aggregating the neighbor information in a temporal graph to generate node representations \cite{DBLP:conf/iclr/XuRKKA20}.
    
    \item  \textbf{TGN} is designed for dynamic graphs, which learns on a sequence of events \cite{DBLP:journals/corr/abs-2006-10637}.

    \item  \textbf{DGSR} explores the dynamic collaborative information between users and items based on different user sequences from the dynamic graph \cite{zhang2022dynamic}.
\end{itemize}

\subsection{Evaluation Metrics}
Three metrics are adopted to evaluate the performance of different models, including Recall, Normalized Discounted Cumulative Gain (NDCG), and Personal Hit Ratio (PHR). 

Recall evaluates the model's ability in selecting all the relevant elements. 
For company $u_i$, Recall is calculated by
\begin{equation}
    \notag
    \mathrm{Recall@K}(u_i) = \frac{|\hat{P}_i \cap P_i|}{|P_i|},
\end{equation}
where $\hat{P}_i$ and $P_i$ are the predicted top-K classification codes and the ground truth of company $u_i$, respectively. $|P|$ denotes the size of a set $P$. The average Recall of all the companies is used as a metric.

NDCG measures the ranking quality by considering the orders of all classification codes. 
For company $u_i$, NDCG is calculated by
\begin{equation}
    \notag
    \mathrm{NDCG@K}(u_i) = \frac{\sum_{k = 1}^{K}{\frac{\delta(\hat{P}_i^k, P_i)}{\log_2(k + 1)}}}{\sum_{k = 1}^{\min(K, |P_i|)}{\frac{1}{\log_2\left(k + 1\right)}}},
\end{equation}
where $\hat{P}_i^k$ denotes the $k$-th predicted classification code of company $u_i$. $\delta\left(v,P\right)$ is 1 when element $v$ is in set $P$, otherwise 0. We calculate the average NDCG of all the companies as a metric. 

PHR evaluates the performance at the company level by calculating the percentage of companies whose predicted classification codes appear in the ground truth at least once. 
PHR is calculated by
\begin{equation}
    \notag
    \mathrm{PHR@K} = \frac{\sum_{i=1}^{N} {\mathds{1}\left(|\hat{P}_i \cap P_i|\right)}}{N},
\end{equation}
where $N$ is the number of testing companies and $\mathds{1}\left(x\right)$ is an indicator function which returns 1 when $x>0$, otherwise 0. 

\subsection{Experimental Settings}
For SVM and RF, we extract the normalized numbers of the lowest-level classification codes as the company's feature. For GRU4Rec, the sequence of patents applied by each company is used as the input. For FPMC and DREAM, we treat patents applied by each company in each year as a set and then feed the sequence of sets into the models. For TGAT, we construct a temporal graph of companies and patent classification codes. For TGN, we sort the patents by time and then utilize the sequence of patents as the input. We search the hidden dimension of all deep learning methods in [64, 128, 256] for the experiment. EDGPAT is trained in a mini-batch manner and is implemented with PyTorch \cite{DBLP:conf/nips/PaszkeGMLBCKLGA19}. 
We set the learning rate to 0.0001 and use Adam \cite{kingma2014adam} as the optimizer. Dropout \cite{DBLP:journals/jmlr/SrivastavaHKSS14} is adopted to prevent over-fitting. We train our model for 300 epochs on all the datasets and employ an early stopping strategy with a patience of 10. We choose the model that achieves the best performance on the validation set for testing. The experiments are conducted on an Ubuntu machine equipped with one Intel(R) Xeon(R) Gold 6130 CPU @ 2.10GHz with 16 physical cores. The GPU device is NVIDIA Tesla T4 with 15 GB memory. The codes and datasets are available at https://github.com/Hope-Rita/EDGPAT.

\begin{table*}[!htbp]
\centering
\caption{Performance of different methods on the different datasets. The best and second-best performances are boldfaced and underlined. We also show the improvements of our method over the best baseline.}
\label{tab:performance_comparison}
\resizebox{2.02\columnwidth}{!}{
\setlength{\tabcolsep}{1.4mm}{

\begin{tabular}{c|c|ccc|ccc|ccc|ccc}
\hline
\multirow{2}{*}{Datasets}   & \multirow{2}{*}{Methods} & \multicolumn{3}{c|}{K=10}                           & \multicolumn{3}{c|}{K=20}                           & \multicolumn{3}{c|}{K=30}                           & \multicolumn{3}{c}{K=40}                            \\ \cline{3-14} 
                            &                          & Recall          & NDCG            & PHR             & Recall          & NDCG            & PHR             & Recall          & NDCG            & PHR             & Recall          & NDCG            & PHR             \\ \hline
\multirow{12}{*}{PHYSICS}   & TOP                      & 0.0531          & 0.0795          & 0.3591          & 0.0833          & 0.0804          & 0.4413          & 0.1142          & 0.0892          & 0.4965          & 0.1295          & 0.0928          & 0.5280          \\
                            & PersonalTOP              & 0.1258          & 0.1764          & 0.5070          & 0.1482          & 0.1686          & 0.5503          & 0.1646          & 0.1688          & 0.5859          & 0.1760          & 0.1699          & 0.6015          \\
                            & SVM                      & 0.1132          & 0.1493          & 0.4681          & 0.1353          & 0.1467          & 0.5123          & 0.1544          & 0.1495          & 0.5536          & 0.1673          & 0.1515          & 0.5736          \\
                            & RF                       & 0.1132          & 0.1490          & 0.4681          & 0.1353          & 0.1466          & 0.5127          & 0.1543          & 0.1494          & 0.5536          & 0.1673          & 0.1514          & 0.5736          \\
                            & FPMC                     & 0.1061          & 0.1533          & 0.4736          & 0.1450          & 0.1544          & 0.5489          & 0.1698          & 0.1579          & 0.5898          & 0.1900          & 0.1622          & 0.6160          \\
                            & DREAM                    & 0.1063          & 0.1568          & 0.4798          & 0.1448          & 0.1571          & 0.5611          & 0.1741          & 0.1621          & 0.6057          & 0.1946          & 0.1669          & 0.6331          \\
                            & GRU4Rec                  & 0.1304          & 0.1821          & 0.5239          & 0.1564          & 0.1792          & 0.5926          & 0.1914          & 0.1832          & 0.6299          & 0.2132          & 0.1880          & 0.6622          \\
                            & TGAT                     & 0.1359          & 0.1872          & 0.5419          & 0.1870          & 0.1899          & 0.6143          & \underline{0.2240}    & 0.1974          & 0.6564          & \underline{0.2524}    & \underline{0.2038}    & 0.6814          \\
                            & TGN                      & 0.1342          & 0.1779          & \underline{0.5526}    & 0.1805          & 0.1824          & \underline{0.6299}    & 0.2184          & 0.1907          & \underline{0.6777}    & 0.2448          & 0.1972          & \underline{0.7027}    \\
                            & DGSR                     & \underline{0.1428}    & \underline{0.1900}    & 0.5440          & \underline{0.1872}    & \underline{0.1956}    & 0.6172          & 0.2104          & \underline{0.1979}    & 0.6458          & 0.2239          & 0.1997          & 0.6585          \\
                            & EDGPAT                   & \textbf{0.1634} & \textbf{0.1968} & \textbf{0.5787} & \textbf{0.2177} & \textbf{0.2067} & \textbf{0.6679} & \textbf{0.2501} & \textbf{0.2131} & \textbf{0.7031} & \textbf{0.2739} & \textbf{0.2190} & \textbf{0.7239} \\
                            & Improvement              & 12.61\%         & 3.46\%          & 4.51\%          & 14.01\%         & 5.37\%          & 5.69\%          & 10.44\%         & 7.13\%          & 3.61\%          & 7.85\%          & 6.94\%          & 2.93\%          \\ \hline
\multirow{12}{*}{MECHANISM} & TOP                      & 0.0410          & 0.0620          & 0.1760          & 0.0731          & 0.0712          & 0.3149          & 0.0967          & 0.0793          & 0.3997          & 0.1260          & 0.0892          & 0.4681          \\
                            & PersonalTOP              & 0.1630          & 0.2060          & \underline{0.5735}    & 0.2099          & 0.2094          & 0.6290          & 0.2279          & 0.2123          & 0.6454          & 0.2381          & 0.2141          & 0.6567          \\
                            & SVM                      & 0.1395          & 0.1643          & 0.5351          & 0.1677          & 0.1654          & 0.5713          & 0.1821          & 0.1684          & 0.5871          & 0.1909          & 0.1701          & 0.6035          \\
                            & RF                       & 0.1393          & 0.1643          & 0.5345          & 0.1676          & 0.1652          & 0.5713          & 0.1819          & 0.1682          & 0.5865          & 0.1910          & 0.1700          & 0.6035          \\
                            & FPMC                     & 0.1614          & 0.2018          & 0.5557          & 0.2146          & 0.2078          & 0.6105          & 0.2440          & 0.2147          & 0.6410          & 0.2648    & 0.2207          & 0.6682          \\
                            & DREAM                    & 0.1350          & 0.1766          & 0.5020          & 0.1923          & 0.1859          & 0.5676          & 0.2310          & 0.1969          & 0.6105          & 0.2613          & 0.2063          & 0.6427          \\
                            & GRU4Rec                  & 0.1644    & 0.2072          & 0.5574          & 0.2278          & 0.2167          & \underline{0.6320}    & 0.2643          & 0.2270          & \underline{0.6648}    & \underline{0.2908}          & 0.2354          & \underline{0.6897}    \\
                            & TGAT                     & \underline{0.1757}          & \underline{0.2238}    & 0.5608          & \underline{0.2430}    & \underline{0.2329}    & 0.6263          & \underline{0.2761}    & \underline{0.2446}    & 0.6620          & \textbf{0.2968} & \underline{0.2540}    & 0.6851          \\
                            & TGN                      & 0.1465          & 0.1954          & 0.5314          & 0.1810          & 0.1942          & 0.5743          & 0.2056          & 0.1993          & 0.6246          & 0.2218          & 0.2034          & 0.6439          \\
                            & DGSR                     & 0.1544          & 0.2063          & 0.5585          & 0.1823          & 0.2002          & 0.6071          & 0.1975          & 0.2012          & 0.6297          & 0.2108          & 0.2042          & 0.6512          \\
                            & EDGPAT                   & \textbf{0.1982} & \textbf{0.2406} & \textbf{0.6348} & \textbf{0.2541} & \textbf{0.2462} & \textbf{0.6778} & \textbf{0.2802} & \textbf{0.2521} & \textbf{0.6993} & \textbf{0.2968} & \textbf{0.2562} & \textbf{0.7111} \\
                            & Improvement              & 11.35\%         & 6.98\%          & 9.66\%          & 4.37\%          & 5.40\%          & 6.76\%          & 1.46\%          & 2.98\%          & 4.93\%          & --              & 0.86\%          & 3.01\%          \\ \hline
\multirow{12}{*}{P \& M}    & TOP                      & 0.0294          & 0.0434          & 0.2428          & 0.0481          & 0.0462          & 0.3062          & 0.0652          & 0.0506          & 0.3593          & 0.0830          & 0.0560          & 0.4071          \\
                            & PersonalTOP              & 0.1254          & 0.1754          & 0.5091          & 0.1556          & 0.1711          & 0.5576          & 0.1694          & 0.1706          & 0.5797          & 0.1761          & 0.1703          & 0.5880          \\
                            & SVM                      & 0.1106          & 0.1420          & 0.4746          & 0.1310          & 0.1390          & 0.5100          & 0.1436          & 0.1401          & 0.5321          & 0.1492          & 0.1401          & 0.5410          \\
                            & RF                       & 0.1105          & 0.1419          & 0.4744          & 0.1309          & 0.1389          & 0.5100          & 0.1435          & 0.1400          & 0.5318          & 0.1491          & 0.1400          & 0.5406          \\
                            & FPMC                     & 0.1072          & 0.1521          & 0.4543          & 0.1439          & 0.1522          & 0.5193          & 0.1669          & 0.1558          & 0.5543          & 0.1840          & 0.1596          & 0.5798          \\
                            & DREAM                    & 0.1143          & 0.1618          & 0.4680          & 0.1568          & 0.1640          & 0.5391          & 0.1865          & 0.1701          & 0.5823          & 0.2092          & 0.1756          & 0.6103          \\
                            & GRU4Rec                  & \underline{0.1321}    & \underline{0.1890}    & 0.5175          & \underline{0.1729}    & \underline{0.1877}    & \underline{0.5830}    & \underline{0.1988}    & \underline{0.1920}    & \underline{0.6156}    & 0.2173          & \underline{0.1962}    & \underline{0.6322}    \\
                            & TGAT                     & 0.1199          & 0.1709          & 0.4957          & 0.1671          & 0.1721          & 0.5662          & 0.1957          & 0.1770          & 0.6062          & \underline{0.2174}    & 0.1824          & 0.6290          \\
                            & TGN                      & 0.1289          & 0.1808          & \underline{0.5299}    & 0.1607          & 0.1770          & 0.5816          & 0.1772          & 0.1775          & 0.5998          & 0.1899          & 0.1795          & 0.6156          \\
                            & DGSR                     & 0.1215          & 0.1699          & 0.5055          & 0.1421          & 0.1622          & 0.5453          & 0.1532          & 0.1611          & 0.5655          & 0.1649          & 0.1627          & 0.5890          \\
                            & EDGPAT                   & \textbf{0.1681} & \textbf{0.2134} & \textbf{0.5945} & \textbf{0.2202} & \textbf{0.2173} & \textbf{0.6648} & \textbf{0.2427} & \textbf{0.2200} & \textbf{0.6885} & \textbf{0.2577} & \textbf{0.2226} & \textbf{0.7000} \\
                            & Improvement              & 21.42\%         & 11.43\%         & 10.87\%         & 21.48\%         & 13.62\%         & 12.30\%         & 18.09\%         & 12.73\%         & 10.59\%         & 15.64\%         & 11.86\%         & 9.69\%          \\ \hline
\multirow{12}{*}{ALL}       & TOP                      & 0.0090          & 0.0225          & 0.1121          & 0.0155          & 0.0217          & 0.1688          & 0.0237          & 0.0237          & 0.2133          & 0.0307          & 0.0260          & 0.2536          \\
                            & PersonalTOP              & \underline{0.0958}    & 0.1601          & \underline{0.4986}    & \underline{0.1247}    & 0.1519          & \underline{0.5561}    & \underline{0.1386}    & 0.1506          & \underline{0.5762}    & 0.1457          & 0.1501          & 0.5843          \\
                            & SVM                      & 0.0406          & 0.1236          & 0.4446          & 0.0610          & 0.1203          & 0.5025          & 0.0737          & 0.1120          & 0.5185          & 0.0825          & 0.1193          & 0.5265          \\
                            & RF                       & 0.0353          & 0.1235          & 0.4152          & 0.0562          & 0.1227          & 0.4830          & 0.0685          & 0.1236          & 0.4970          & 0.0776          & 0.1236          & 0.5070          \\
                            & FPMC                     & 0.0624          & 0.1167          & 0.3570          & 0.0858          & 0.1102          & 0.4133          & 0.1004          & 0.1101          & 0.4499          & 0.1112          & 0.1115          & 0.4747          \\
                            & DREAM                    & 0.0729          & 0.1292          & 0.4114          & 0.1025          & 0.1251          & 0.4823          & 0.1231          & 0.1274          & 0.5259          & 0.1392          & 0.1308          & 0.5545          \\
                            & GRU4Rec                  & 0.0921          & \underline{0.1675}    & 0.4855          & 0.1202          & \underline{0.1563}    & 0.5414          & 0.1382          & \underline{0.1557}    & 0.5709          & \underline{0.1516}    & \underline{0.1573}    & \underline{0.5937}    \\
                            & TGAT                     & 0.0601          & 0.1032          & 0.3728          & 0.0892          & 0.1022          & 0.4517          & 0.1087          & 0.1055          & 0.4961          & 0.1246          & 0.1095          & 0.5285          \\
                            & TGN                      & 0.0828          & 0.1391          & 0.4604          & 0.1017          & 0.1300          & 0.5033          & 0.1114          & 0.1279          & 0.5294          & 0.1195          & 0.1281          & 0.5482          \\
                            & DGSR                     & 0.0832          & 0.1289          & 0.4537          & 0.1038          & 0.1230          & 0.4960          & 0.1121          & 0.1213          & 0.5059          & 0.1165          & 0.1206          & 0.5113          \\
                            & EDGPAT                   & \textbf{0.1175} & \textbf{0.1725} & \textbf{0.5491} & \textbf{0.1646} & \textbf{0.1742} & \textbf{0.6304} & \textbf{0.1868} & \textbf{0.1769} & \textbf{0.6612} & \textbf{0.2006} & \textbf{0.1769} & \textbf{0.6800} \\
                            & Improvement              & 18.47\%         & 2.90\%          & 9.20\%          & 24.24\%         & 10.28\%         & 11.79\%         & 25.80\%         & 11.98\%         & 12.86\%         & 24.43\%         & 11.08\%         & 12.69\%         \\ \hline
\end{tabular}

}}
\end{table*}
\subsection{Performance Comparison}
\label{subsec:performance_all_patents}

The comparisons of all the methods on top-K performance on different datasets are shown in \tabref{tab:performance_comparison}. From \tabref{tab:performance_comparison}, several conclusions can be summarized as follows.
Firstly, PersonalTOP achieves much better performance than TOP because it focuses on the personalized sequence for prediction while TOP solely provides identical predictions for all the companies. The results indicate that different companies tend to focus on different technical directions, and therefore individual information is essential in patent application trend prediction. 

Secondly, since the inputs of SVM and RF also contain personalized information, they obtain much better performance than TOP, which again validates the importance of the company's personalized characteristics.

Thirdly, DREAM and GRU4Rec achieve better performance than FPMC in most cases, which demonstrates the effectiveness of capturing sequential dependencies with RNNs while FPMC just models the adjacent information with the Markov chain. Therefore, learning the whole sequence is beneficial for capturing the dynamics of companies and classification codes. Moreover, as the number of fields increases, GRU4Rec shows superior performance in DREAM and FPMC because it learns on the sequence in a fine granularity and could capture more comprehensive temporal information.

Fourthly, TGAT or DGSR performs better than other baselines in PHYSICS and MECHANISM in most cases, which illustrates that capturing high-order temporal patterns by graph convolutions is beneficial. However, the performance of TGAT and DGSR decreases rapidly with datasets involving more fields. This is because the relationships between classification codes are more complex when the number of fields increases and TGAT fails to capture their semantic correlations. Besides, TGN shows better performance than TGAT in most cases on P \& M and ALL, because TGN sequentially combines the current information and historical states, which is more effective than the single aggregating process in TGAT. Besides, we find that sometimes the NDCG may decrease in some models with the increment of K when K is small. This may be caused by the increment of labels that are correctly classified being smaller than its increment of K in the denominator. However, when K is enough large, the NDCG will increase with the increment of K.

Finally, EDGPAT significantly outperforms the existing methods due to: 1) the learning of event-based continuous-time representations of both companies and patent classification codes, which well mine their evolving patterns; 2) the explorations of classification code semantics based on the hierarchical taxonomy; 3) the combination of representations of companies and classification codes from the static, dynamic and hierarchical perspectives.

\subsection{Robustness to the Ratio of Training Data}
\begin{figure}[!htbp]
    \centering
    \includegraphics[width=1.0\columnwidth]{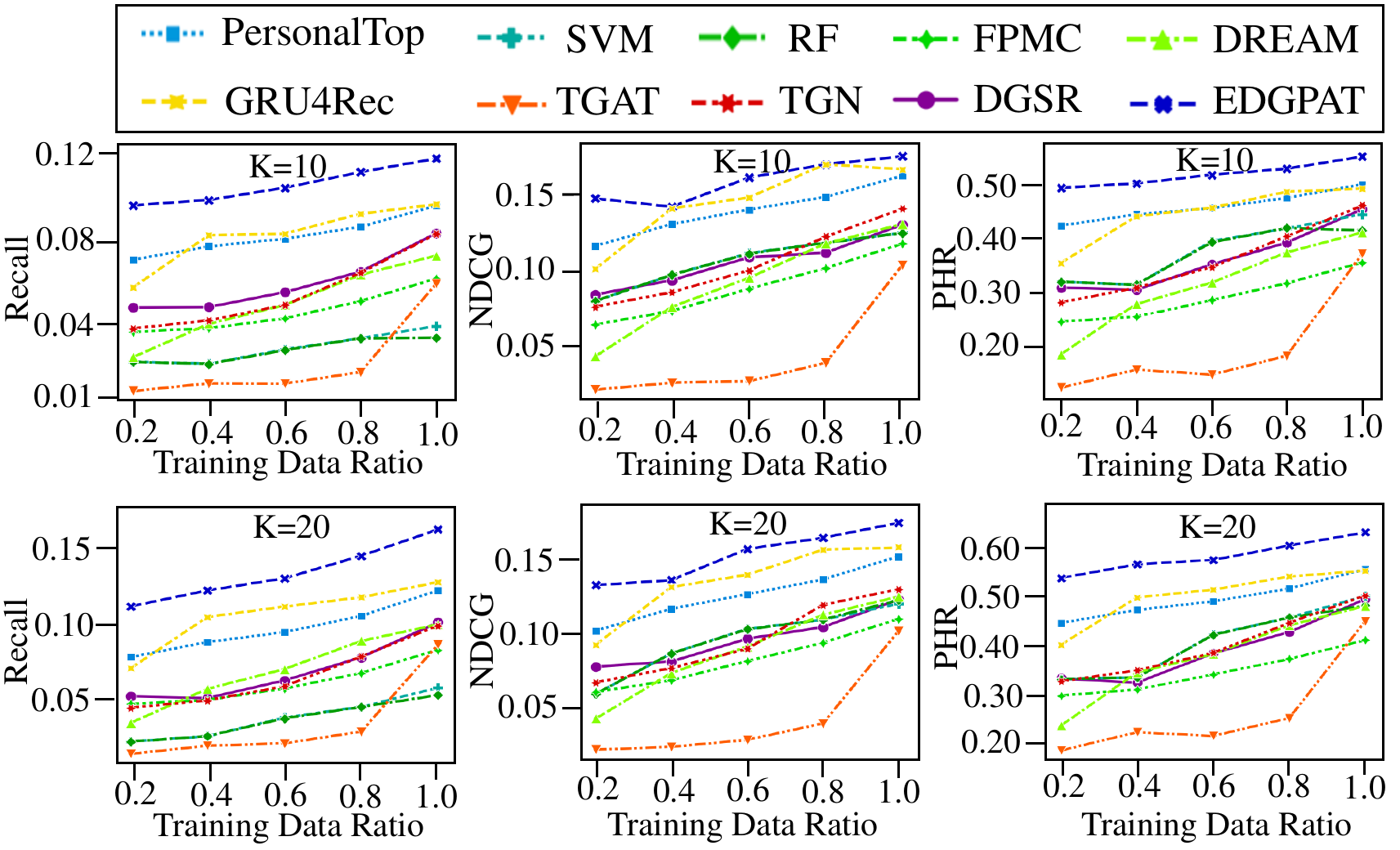}
    \caption{Performance of different methods with different ratios of the training data, where the values of K are set to 10 and 20. TOP is not shown due to its inferior performance. When the values of K are set to 30 and 40, we obtain similar observations and thus do not plot them due to space limitations. }
    \label{fig:data_ratio}
\end{figure}

As EDGPAT maintains and continuously updates the memorable representations of companies and patent classification codes when a patent is observed, it can utilize the data more sufficiently. This may provide EDGPAT with the ability to achieve satisfactory performance with a part of the training data. To this end, we validate the robustness of EDGPAT by varying the ratio of training data in [20\%, 40\%, 60\%, 80\%, 100\%] on the ALL dataset. The performance of all the methods is shown in \figref{fig:data_ratio}.

From \figref{fig:data_ratio}, we conclude that the performance of EDGPAT is more stable to training data ratios than the baselines, and it achieves the best performance even with a portion of the training data than baselines that use all the training data. This demonstrates the advantage of EDGPAT in comprehensively using the training data by computing each patent in chronological order.

\subsection{Incorporating Patent Texts into Models}
Our approach predicts the patent application trend by solely using the patent application records of companies. In practice, a patent is also associated with text descriptions, which explicitly indicate the patent semantics. Therefore, in this subsection, we aim to incorporate the patent texts into models to show the influence of texts on prediction performance. We compare our method with GRU4Rec and TGN since these two methods achieve relatively better results in the above experiments.

\begin{figure}[!htbp]
    \centering
    \includegraphics[width=1.\columnwidth]{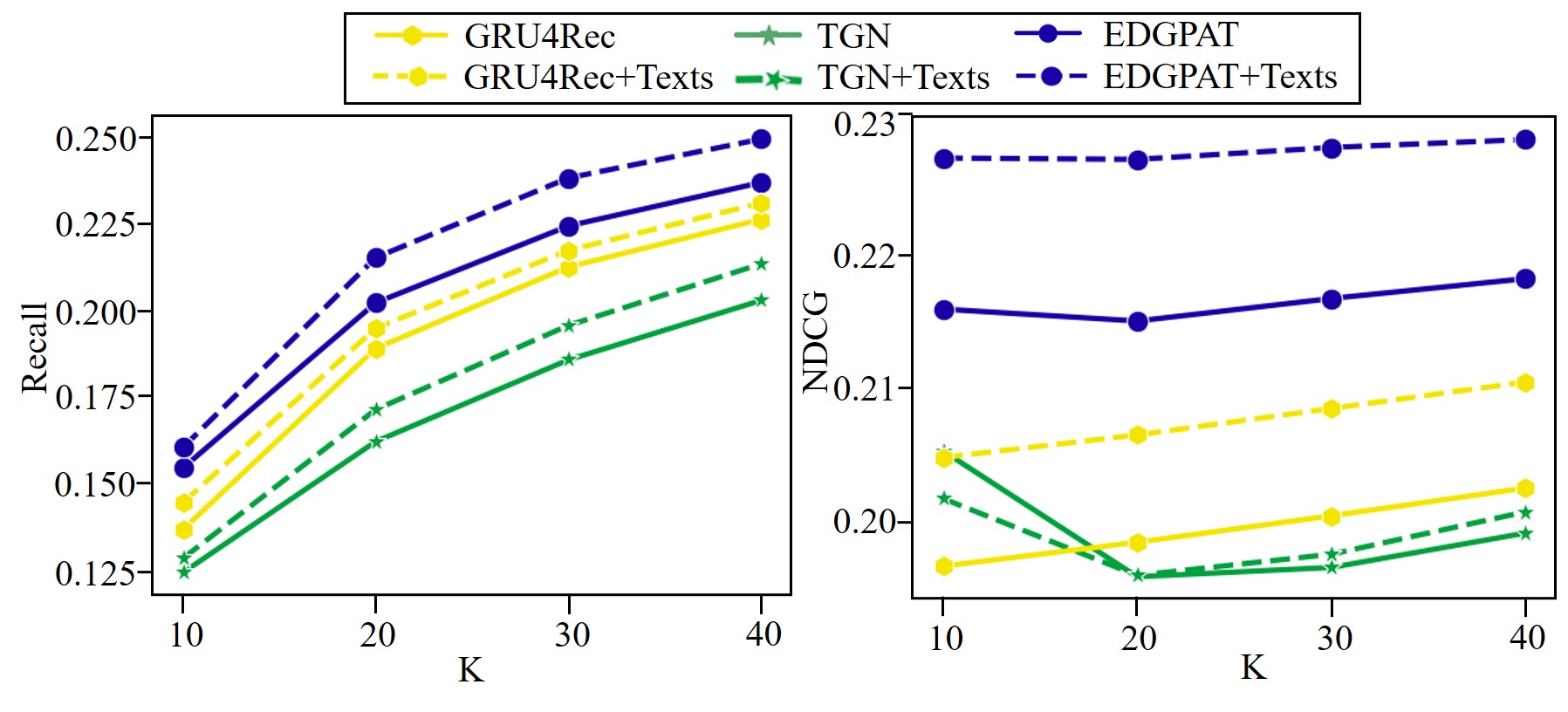}
    \caption{Performance of different methods with textual embeddings.}
    \label{fig:addText embeddings}
\end{figure}

In particular, we first select patents that have text descriptions in the Summary part and remove words appearing less than 5 times for text corpus construction. Then, we apply Gensim \textit{Word2vec} tool \footnote{https://radimrehurek.com/gensim/models/word2vec.html} to train the 100-dimensional word embedding for each word in the text corpus. Next, We encode the texts with Bi-LSTM \cite{DBLP:journals/tsp/SchusterP97} to obtain the vectorized textual representation for each patent. Finally, we inject the textual representation into models by concatenating it with the original inputs. Experimental results are shown in \figref{fig:addText embeddings}, where the dashed/solid lines represent the performance of models with/without textual representations. PHR is not plotted due to space limitations.



From \figref{fig:addText embeddings}, we conclude that all the models often perform better when considering the texts of patents, indicating the semantic information of patents can contribute to the prediction of patent application trends. Moreover, whether or not the patent semantics are considered, the proposed EDGPAT can consistently outperform GRU4Rec and TGN, which demonstrates the good generality of our model.

\subsection{Performance on Patent Application Trend Prediction at Higher Levels }
\begin{figure}[!htbp]
    \centering
    \includegraphics[width=1.\columnwidth]{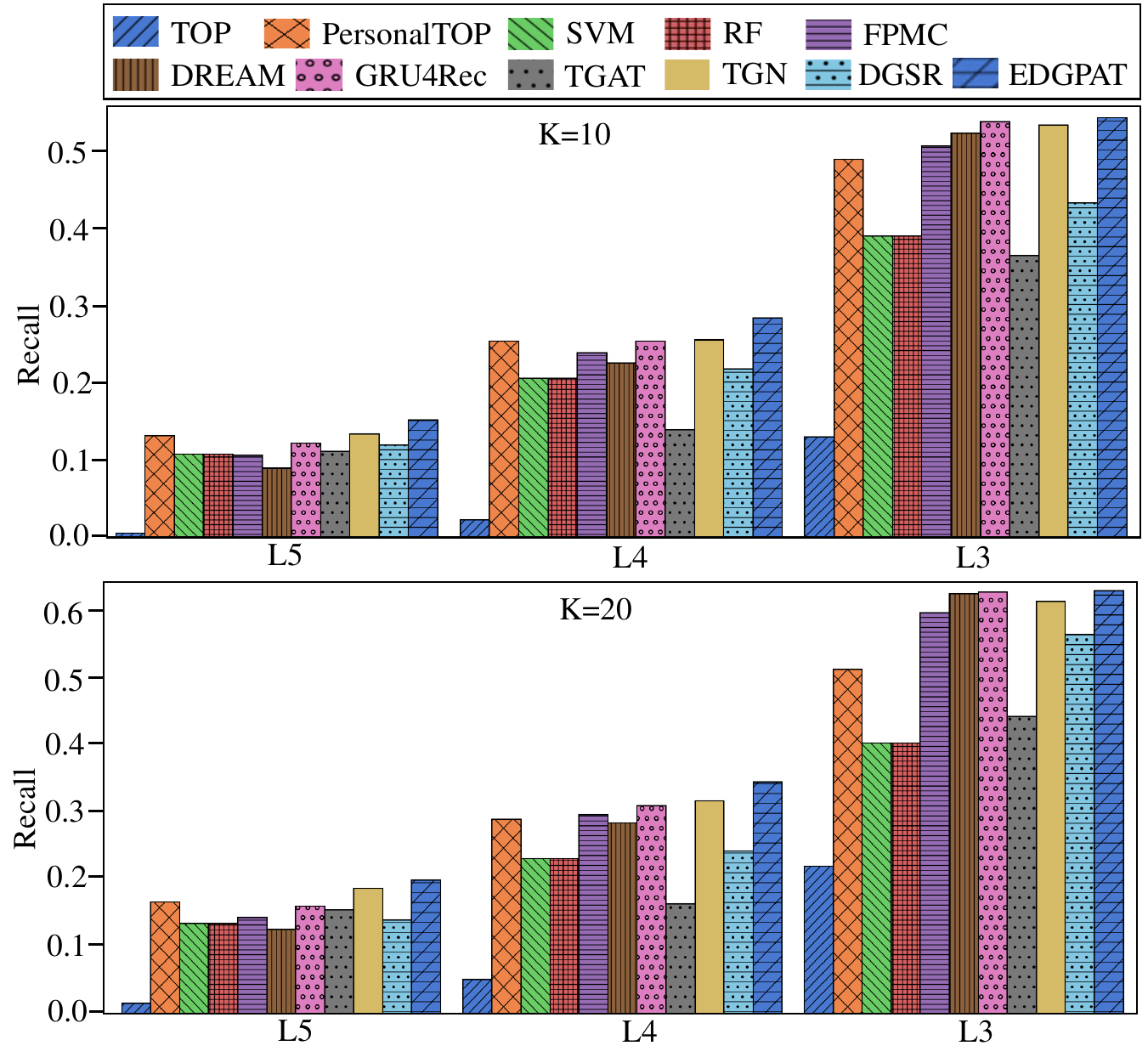}
    \caption{Performance on Application Trend Prediction at Different Levels.}
    \label{fig:different levels}
\end{figure}
In addition to predicting the lowest-level (i.e., 5th-level) classification codes to infer the patent application trend, we also validate the model performance when predicting classification codes at higher levels, including the 3rd-level and the 4-th level. Performance of all the methods on the Recall metric is shown in \figref{fig:different levels}. NDCG and PHR are not reported due to space limitations.

From \figref{fig:different levels}, we could find that the performance becomes better when making predictions at higher levels since the numbers of classification codes are less at higher levels and make the task easier. Moreover, EDGPAT is still able to outperform the baselines due to the capturing of semantic correlations of classification codes.


\subsection{Ablation Study}
\label{subsec:ablation std}
\begin{figure}[!htbp]
    \centering
    \includegraphics[width=1\columnwidth]{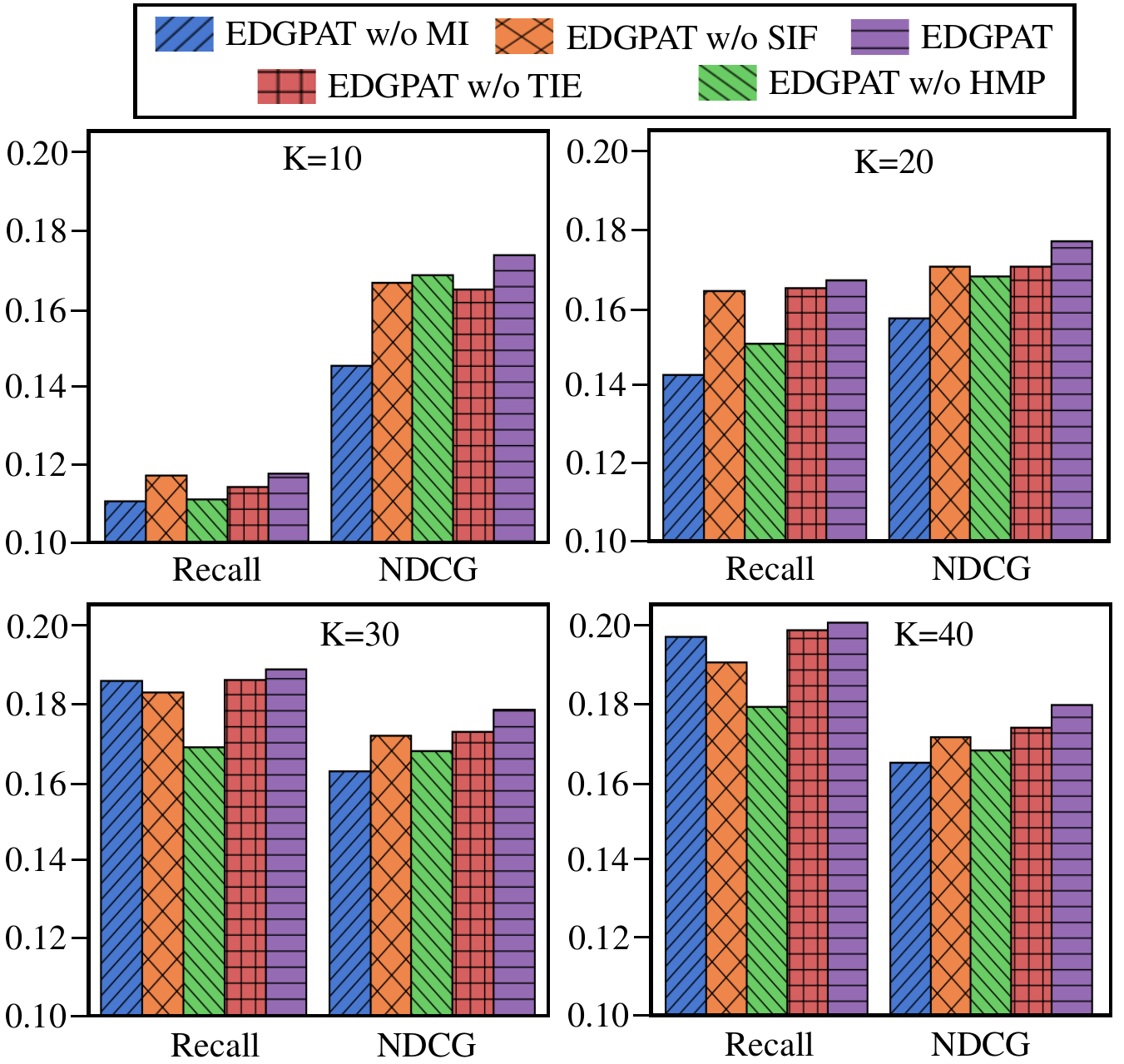}
    \caption{Effects of different components in EDGPAT.}
    \label{fig:ablation_study}
\end{figure}

We validate the effectiveness of the Message Interactions in the encoding process (MI), the Static Information Fusion (SIF), the Hierarchical Message Passing (HMP), and Temporal Information Encoding (TIE) by removing these four components and then comparing their performance with EDGPAT. 
Specifically, EDGPAT w/o MI computes the messages for companies and classification codes solely based on their messages. EDGPAT w/o SIF removes the static information in multi-perspective representations fusing module for prediction. EDGPAT w/o HMP does not use the hierarchical taxonomy. EDGPAT w/o TIE is implemented by removing temporal information in the message encoders. 
We show the performance of Recall and NDCG metrics when the values of K are set to 10, 20, 30, and 40 in \figref{fig:ablation_study}. Similar observations could be found on the PHR.

From \figref{fig:ablation_study}, we could find that EDGPAT achieves the best performance when it is equipped with all components, and the results would decrease when any component is abandoned. 
Specifically, message interactions in the encoding process help companies and classification codes communicate with each other. Static information could reflect the stationary characteristics of companies and classification codes. The hierarchical message passing component explicitly captures the semantic correlations of classification codes. Besides, we observe that Message Interactions information plays an important role when K is set as 10 and 20 and the Hierarchical Message Passing module is more important when K is set as 30 and 40. It is worth noting that our model could still perform better than baselines even without using the hierarchical taxonomy.


\subsection{Model Efficiency Comparison} 
\begin{figure}[!htbp]
    \centering
    \includegraphics[width=1.0\columnwidth]{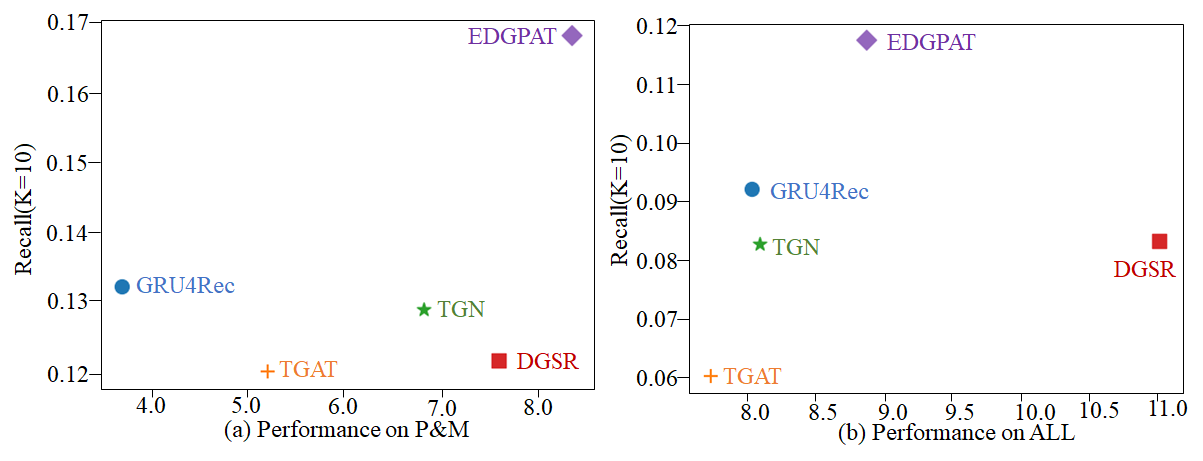}
    \caption{Log-scale evaluation time of different methods on P\&M and ALL datasets.}
    \label{fig:runtime_comparison}
\end{figure}

We further present the evaluation time of different methods P\&M and ALL datasets in \figref{fig:runtime_comparison}, where the x-axis is scaled by a logarithmic function with base 2. The model complexity of EDGPAT mostly comes from the Event-based Continuous-time Representation Learning and Hierarchical Message Passing components, which is relatively higher than other baselines. However, we also observe that EDGPAT obtains the best performance with acceptable increments in computational complexity. Therefore, we conclude that EDGPAT can achieve a good trade-off between efficiency and effectiveness. 

\subsection{Model Interpretability}
We also show the interpretability of our approach. 

\begin{figure}[!htbp]
    \centering
    \includegraphics[width=0.8\columnwidth]{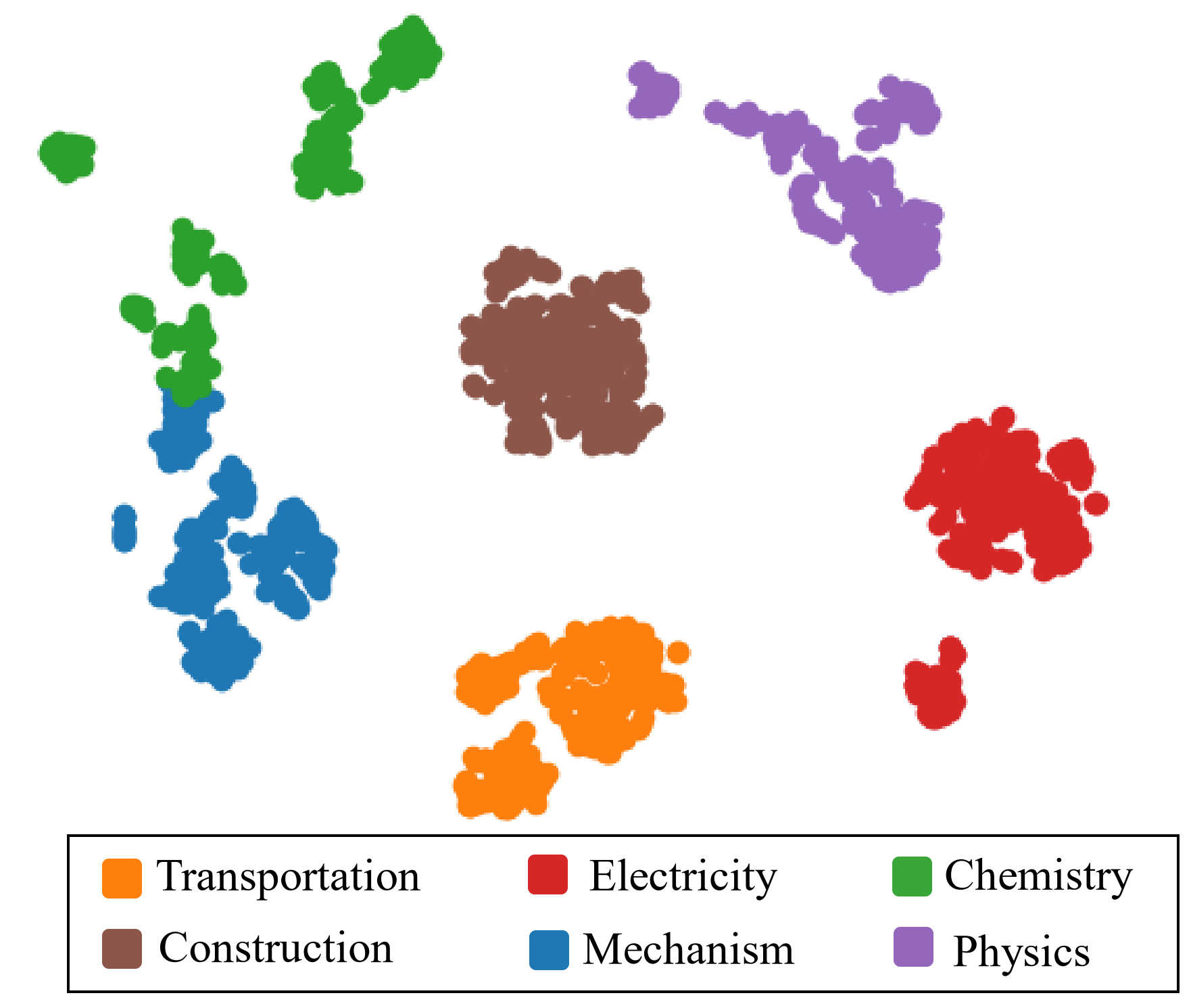}
    \caption{Visualization of classification codes in different research fields.}
    \label{fig:field_visualization}
\end{figure}

\textbf{Classification Codes Visualization}.
We first randomly sample six types of classification codes from different research fields and set the number of classification codes in each research field to 500. Then, we visualize the memorable representations of the classification codes with t-SNE \cite{tNSE} in \figref{fig:field_visualization}. We could find that: 1) classification codes in the same research field are gathered closely and the boundaries between classification codes with different research fields are more prominent. This demonstrates the effectiveness of our method in capturing semantic proximities of classification codes; 2) there also exist tiny clusters of classification codes in the same research field, which may be caused by the unique characteristics of some classification nodes.

\begin{figure}[!htbp]
    \centering
    \includegraphics[width=1.0\columnwidth]{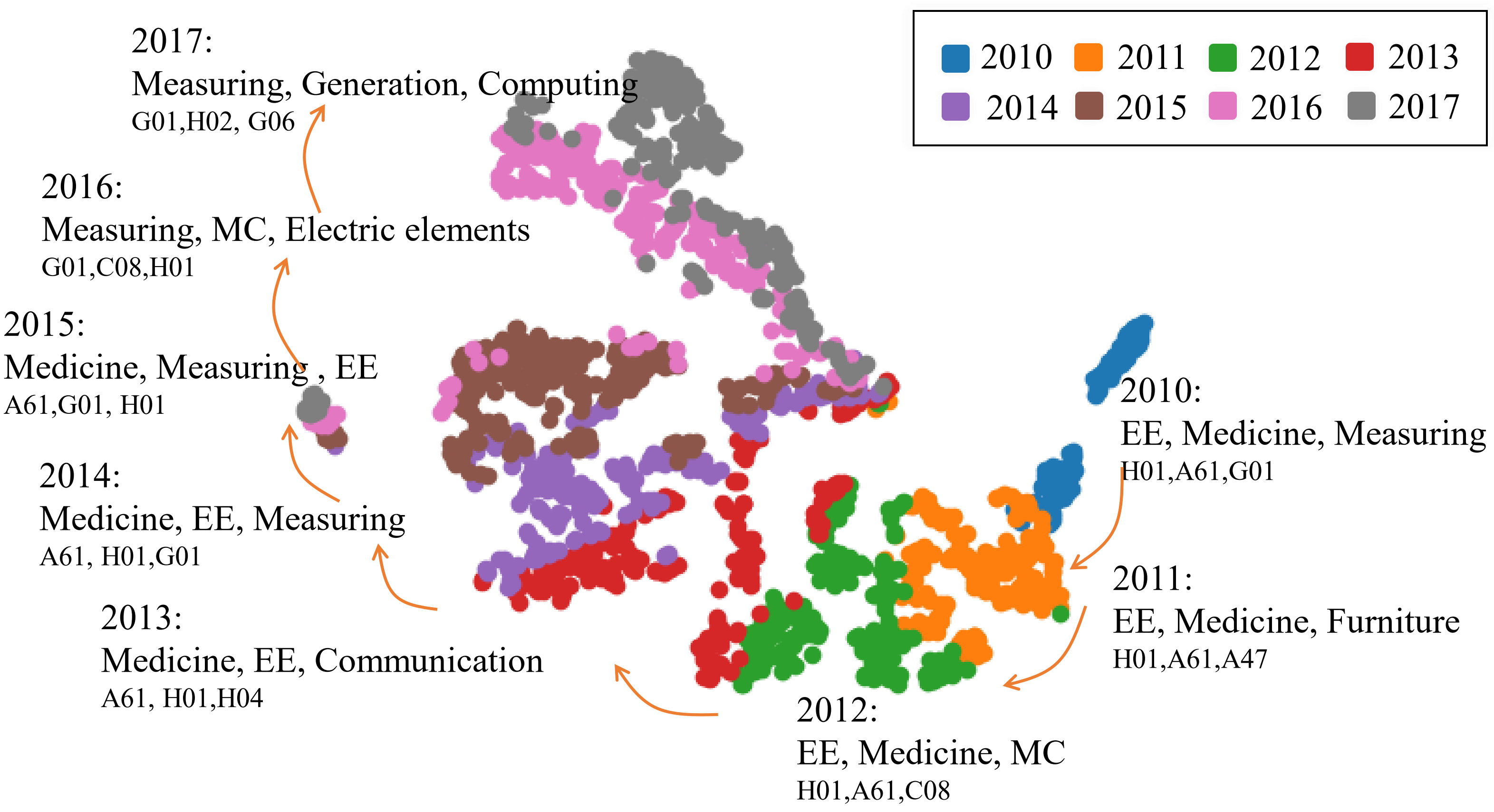}
    \caption{Visualization of the memorable representations of a randomly sampled company over time. Top-3 research fields and the corresponding classification codes at the $4$-th level are shown. EE and MC are the abbreviations of Electric Elements and Macromolecular Compounds.}
    \label{fig:company_visualization}
\end{figure}

\textbf{Company Developing Trajectory Visualization}. 
We randomly select a company and project all of its memorable representations with t-SNE \cite{tNSE} in \figref{fig:company_visualization}. From \figref{fig:company_visualization}, we could observe that the company state is evolving over time with the developing trends in different research fields. Our approach can well capture the company's developing trajectory by learning the continuously evolving representations of the company, demonstrating the good ability of our model for company analysis.

\section{Conclusion}
\label{section-6}
This paper formalized a new task: predicting what types of patents a company will apply for in the next period. To cope with the problem, a dynamic graph representation learning framework was proposed, which was founded on learning memorable representations of companies and classification codes. In particular, our method consists of three components: 1) an event-based continuous-time representation learning module to maintain and update the memories of companies and classification codes; 2) a hierarchical message passing component to learn the semantic correlations of classification codes; and 3) a multi-perspective information fusing module to aggregate the static, dynamic and hierarchical representations. Experimental results on real-world datasets showed that our method could achieve better performance than the baselines in a variety of settings. The abilities of our approach in learning the semantic correlations of classification codes and tracking the development trajectories of companies are also demonstrated.


%

%

\ifCLASSOPTIONcompsoc
  \section*{Acknowledgments}
\else
  \section*{Acknowledgment}
\fi
This work was supported by the National Key R\&D Program of China (2021YFB2104802), the National Natural Science Foundation of China (62272023, 62176014, 62276015) and the Fundamental Research Funds for the Central Universities (No. YWF-23-L-1203).

\ifCLASSOPTIONcaptionsoff
  \newpage
\fi



\bibliographystyle{IEEEtran}
\bibliography{reference}

%
\begin{IEEEbiography}[{\includegraphics[width=1in,height=1.25in,clip,keepaspectratio]{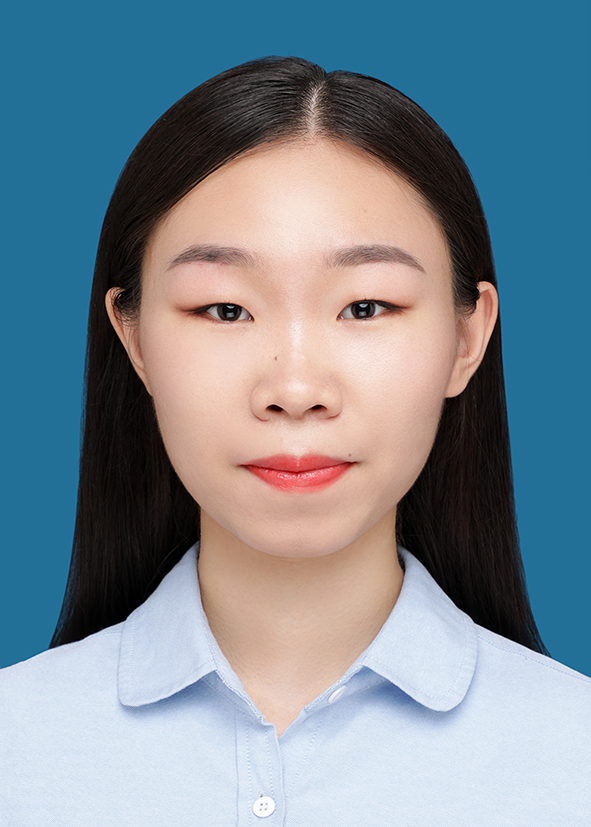}}]
{Tao Zou} received the B.S. degree from the School of Computer Science and Engineering, Beihang University, China, in 2021. She is currently pursuing the  M.S. degree in the School of Computer Science and Engineering, Beihang University, China. Her research interests include dynamic graph learning, machine learning and knowledge data mining.
\end{IEEEbiography}

\begin{IEEEbiography}[{\includegraphics[width=1in,height=1.25in,clip,keepaspectratio]{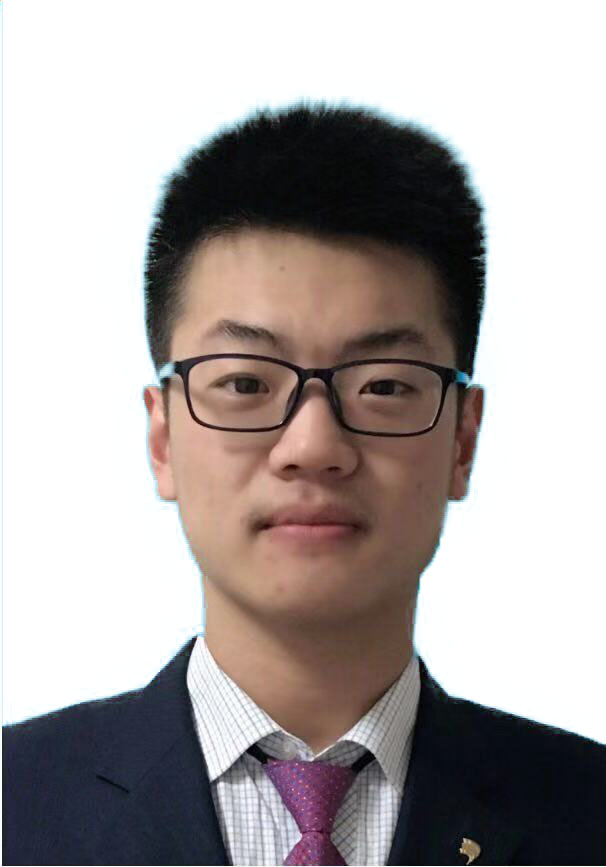}}]
{Le Yu} received the B.S. degree in Computer Science and Engineering from Beihang University, Beijing, China, in 2019. He is currently a third-year computer science Ph.D. student in the School of Computer Science and Engineering at Beihang University. His research interests include temporal data mining, machine learning and graph neural networks.
\end{IEEEbiography}

\begin{IEEEbiography}[{\includegraphics[width=1in,height=1.25in,clip,keepaspectratio]{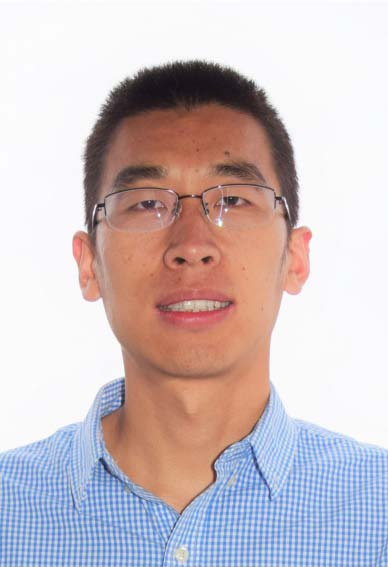}}]
{Leilei Sun} is currently an assistant professor in School of Computer Science, Beihang University, Beijing, China. He was a postdoctoral research fellow from 2017 to 2019 in School of Economics and Management, Tsinghua University. He received his Ph.D. degree from the Institute of Systems Engineering, Dalian University of Technology, in 2017. His research interests include machine learning and data mining. 
\end{IEEEbiography}

\begin{IEEEbiography}[{\includegraphics[width=1in,height=1.25in,clip,keepaspectratio]{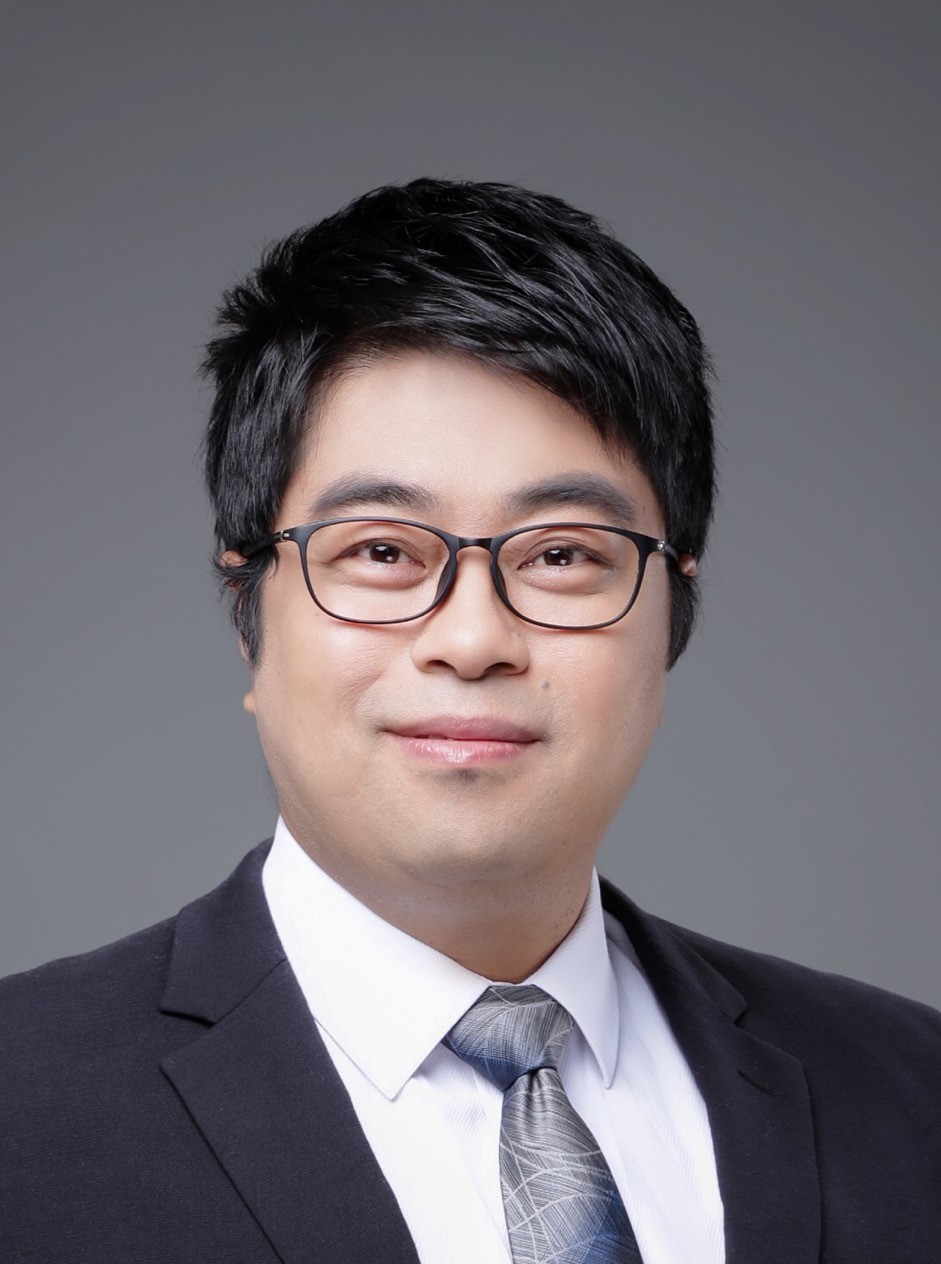}}]
{Bowen Du} received the Ph.D. degree in Computer Science and Engineering from Beihang University, Beijing, China, in 2013. He is currently a Professor with the State Key Laboratory of Software Development Environment, Beihang University. His research interests include smart city technology, multi-source data fusion, and traffic data mining.
\end{IEEEbiography}

\begin{IEEEbiography}[{\includegraphics[width=1in,height=1.25in,clip,keepaspectratio]{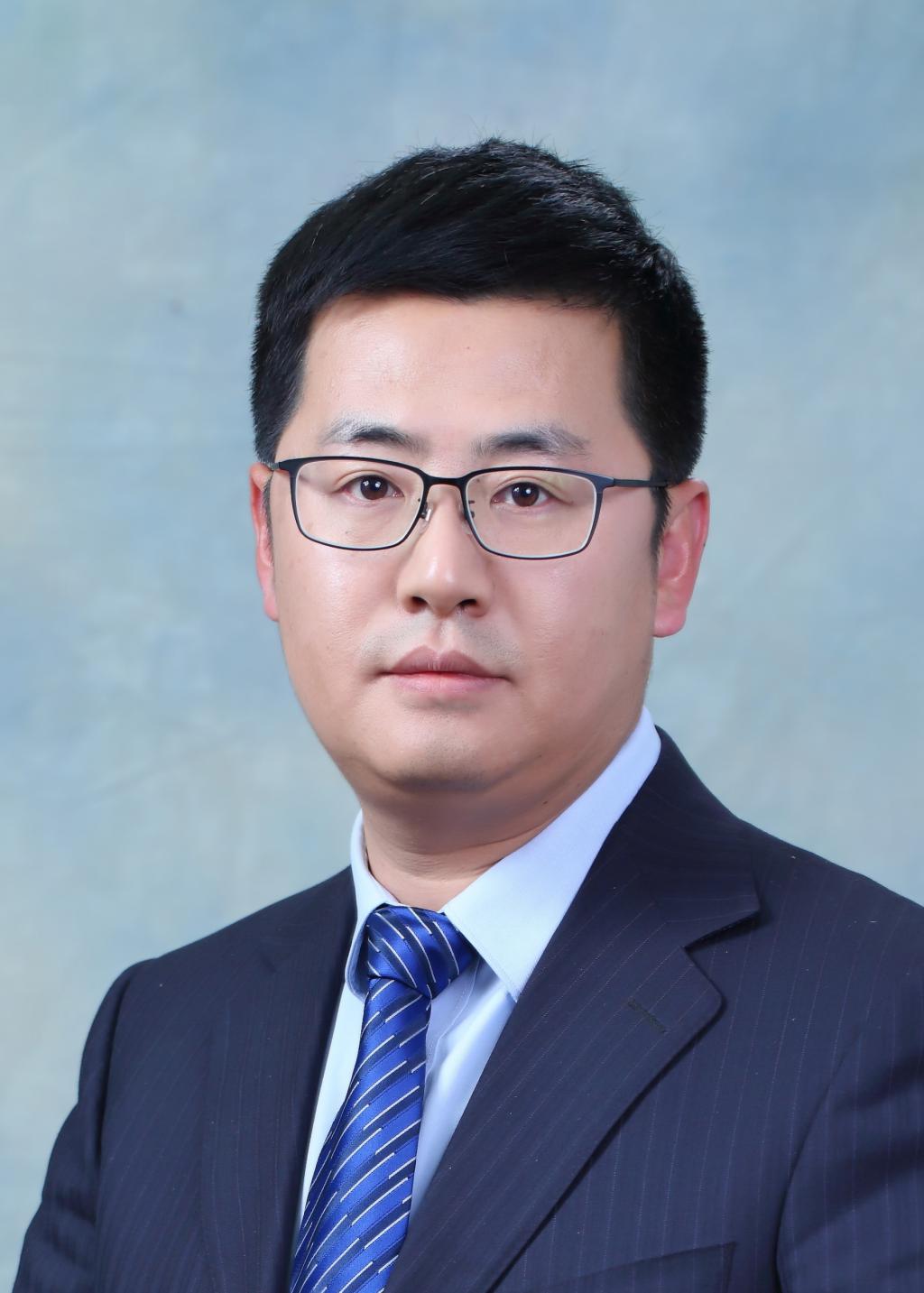}}]
{Deqing Wang} received the Ph.D. degree in computer science from Beihang University, Beijing, China, in 2013. He is currently an Associate Professor at the School of Computer Science and the Deputy Chief Engineer with the National Engineering Research Center for Science Technology Resources Sharing and Service, Beihang University. His research focuses on text categorization and data mining for software engineering and machine learning.
\end{IEEEbiography}

\begin{IEEEbiography}[{\includegraphics[width=1in,height=1.25in,clip,keepaspectratio]{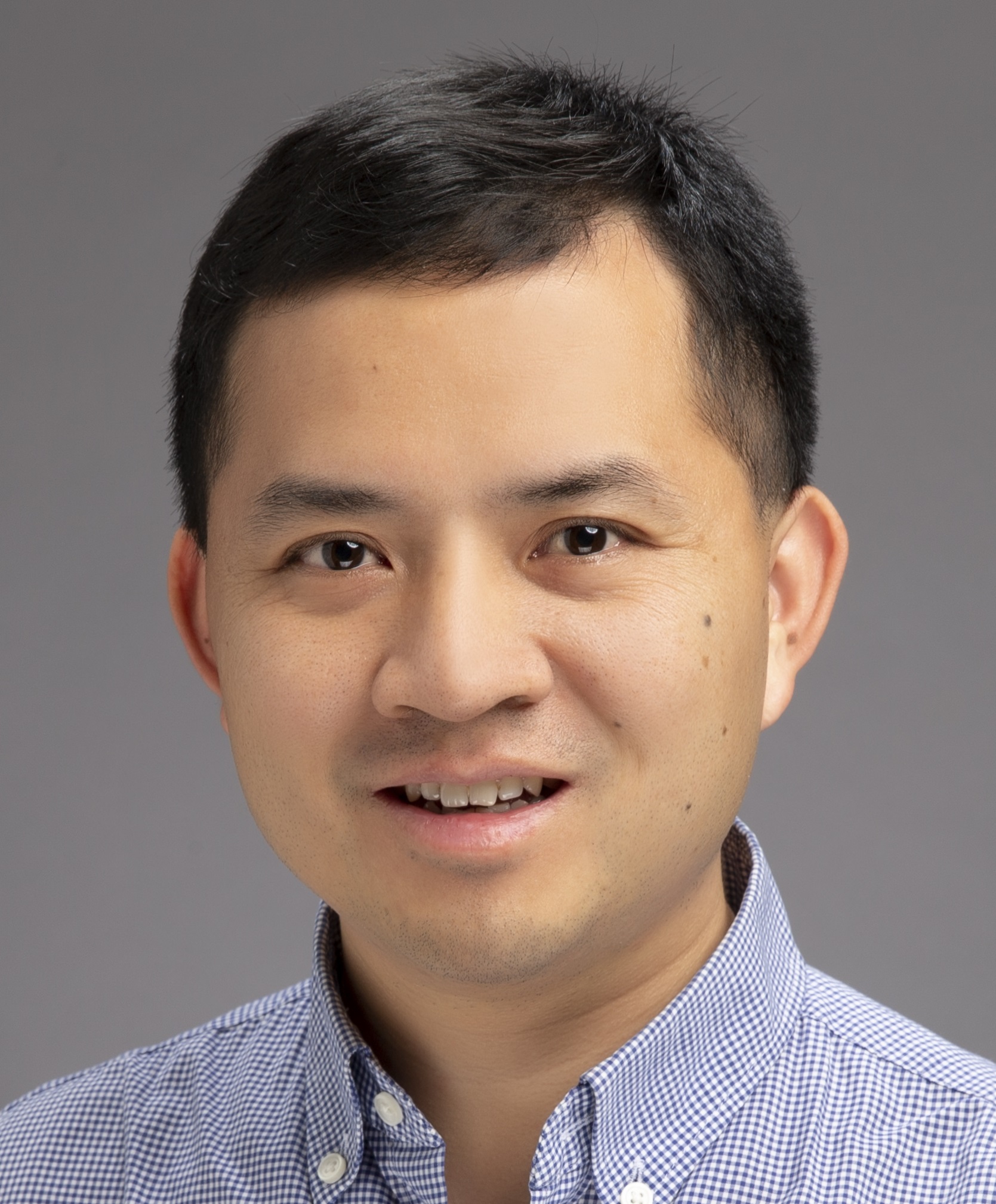}}]
{Fuzhen Zhuang} is currently a Professor in Institute of Artificial Intelligence, Beihang University, Beijing, China. He received the Ph.D. degrees in computer science from the Institute of Computing Technology, Chinese Academy of Sciences, Beijing, China, in 2011. His research interests include machine learning, data mining, transfer learning, multi-task learning, recommendation systems and knowledge graph. He has published over 150 papers in the prestigious refereed journals and conference proceedings, such as Nature Communications, IEEE TKDE, Proc. of IEEE, TNNLS, TIST, KDD, WWW, SIGIR, NeurIPS, IJCAI, AAAI, and ICDE.
\end{IEEEbiography}

\end{document}